\documentclass[11pt]{article}
\usepackage[a4paper]{geometry}

\usepackage{natbib}

\usepackage[utf8]{inputenc} 
\usepackage[T1]{fontenc}    
\usepackage[colorlinks]{hyperref}        
\usepackage{url}            
\usepackage{booktabs}       
\usepackage{amsfonts}       
\usepackage{nicefrac}       
\usepackage{microtype}      
\usepackage[usenames,dvipsnames]{xcolor}
\usepackage{algorithm}
\usepackage{algorithmic}
\usepackage{graphicx,epsfig,tikz,caption}
\usepackage{bm}

\usepackage{lmodern}
\usepackage{times}
\usepackage{mathtools}

\usepackage{amssymb,amsmath,amsthm}
\usepackage{bbold}
\usepackage{comment}
\usepackage{subcaption}

\usepackage{parskip}

\definecolor{BrickRed}{rgb}{0.6,0,0}
\definecolor{RoyalBlue}{rgb}{0,0,0.8}
\definecolor{Tdgreen}{rgb}{0,0.4,0.7}
\hypersetup{citecolor=MidnightBlue, linkcolor=BrickRed}

\newtheorem{theorem}{Theorem}

\newtheorem{lemma}{Lemma}
\newtheorem{proposition}{Proposition}

\DeclareMathOperator*{\argmax}{argmax}

\def\RR{{\mathbb R}}

\newcommand{\A}{\mathcal{A}}

\newcommand*{\affmark}[1][*]{\textsuperscript{#1}}
\newcommand*{\email}[1]{\texttt{#1}}

\title{\textbf{Contextual Linear Bandits under Noisy Features: \\ Towards Bayesian Oracles}}

\author{\normalsize \hspace{10mm}
\begin{tabular}{c} \textbf{Jung-hun Kim}\affmark[1] \\ KAIST  \end{tabular} \and
\normalsize\begin{tabular}{c} \textbf{Se-Young Yun}\affmark[1]  \\ KAIST  \end{tabular} \and
\normalsize\begin{tabular}{c} \textbf{Minchan Jeong}\affmark[1]  \\ KAIST  \end{tabular} \and
\normalsize \hspace{10mm}
\begin{tabular}{c} \textbf{Junhyun Nam}\affmark[1]  \\ KAIST  \end{tabular} \and
\normalsize\begin{tabular}{c} \textbf{Jinwoo Shin}\affmark[1]  \\ KAIST \end{tabular} \and
\normalsize\begin{tabular}{c} \textbf{Richard Combes}\affmark[2]   \\ Centrale-Supelec \end{tabular} \and
 \normalsize\email{\affmark[1]\{junghunkim, yunseyoung, mcjeong, junhyun.nam, jinwoos\}@kaist.ac.kr}\\
 \normalsize\email{\affmark[2]richard.combes@centralesupelec.fr}
}

\date{}

\begin{document}

\maketitle

\begin{abstract}
  We study contextual linear bandit problems under feature uncertainty, where the features are noisy and have missing entries. To address the challenges posed by this noise, we analyze Bayesian oracles given the observed noisy features. Our Bayesian analysis reveals that the optimal hypothesis can significantly deviate from the underlying realizability function, depending on the noise characteristics. These deviations are highly non-intuitive and do not occur in classical noiseless setups. This implies that classical approaches cannot guarantee a non-trivial regret bound. Therefore, we propose an algorithm that aims to approximate the Bayesian oracle based on the observed information under this model, achieving $\tilde{O}(d\sqrt{T})$ regret bound when there is a large number of arms. We demonstrate the proposed algorithm using synthetic and real-world datasets.
\end{abstract}

\section{INTRODUCTION} \label{sec:intro}
The bandit problem \citep{lai1985asymptotically} is a fundamental sequential decision-making problem when dealing with the exploration-exploitation trade-off. It has received considerable attention due to its applicability to a wide range of real-world problems, such as clinical trials \citep{thompson1933likelihood}, economics \citep{schlag1998imitate}, routing \citep{awerbuch2004adaptive}, and ranking \citep{radlinski2008learning}.
 In a basic multi-armed-bandit (MAB) problem, there is a finite number of actions or ``arms'' and in each round, an agent selects an arm and observes a random reward. The goal is to minimize regret, which is the difference in expected cumulative reward between the agent and an oracle policy that knows latent parameters. 
 
  A natural extension for the basic MAB is to provide the agent with contextual information \citep{langford2008epoch} that 
  is present in many real-life problems such as personalized recommendations \citep{bouneffouf2012contextual}, web server defense \citep{jung2012contextual}, and information retrieval \citep{hofmann2011contextual}. In each round, the environment draws a context, and the agent observes it. Then the agent chooses an arm based on the contextual information and receives a random reward. contextual linear bandit problems include feature maps between context and arms, so each arm has a feature vector in $\mathbb{R}^d$. There is also a latent parameter in $\mathbb{R}^d$, and the mean reward for each arm follows a linear model between the latent parameter and feature vector. For contextual linear bandit problems, \citet{auer2002using,chu2011contextual}, and \citet{abbasi2011improved} proposed algorithms based on the principle of optimism in the face of uncertainty.

 The uncertainty of features is an important issue for many domains, including computational biology, clinical studies, and economics \citep{sterne2009multiple,troyanskaya2001missing,wooldridge2007inverse}. Therefore, the estimation of latent parameters or learning models under noisy observations has been widely studied by \citet{loh2011high,lounici2014high,pavez2020covariance}, and \citet{you2020handling}. Recommendation systems \citep{li2010contextual,balakrishnan2018using} can construct feature vectors of items by pre-processing item information  such as text descriptions, categories, or figures; it is natural to have some feature noise from the pre-processing. Noisy features can even have missing entries for several reasons, including communication failure and human error. Noise can also be intentionally added to features to preserve privacy. For example, features often represent user profile information, and a recent trend in providing services that respect privacy (called differential privacy \citep{dwork2008}) is to add noise to user profile information. 
Therefore, noisy feature information seems natural and essential for real-world applications.

   In this paper, we consider a variant of the contextual linear bandit problem in which random noise exists in the feature vectors. 
   Here, we briefly describe these noisy features. At each time $t$, the true feature $z_{a,t}\in \mathbb{R}^d$ for an arm $a$ in the set of arms is generated randomly encoding context, and the mean reward is $z_{a,t}^\top \theta^\star$ where $\theta^\star\in\mathbb{R}^d$ is a latent parameter. An agent can only observe a noise feature vector $x_{a,t}\in \mathbb{R}^d$, rather than $z_{a,t}$, which is defined as $x_{a,t}=(z_{a,t}+\varepsilon_{a,t})\circ m_{a,t}$ where $\circ$ is the element-wise product, $\varepsilon_{a,t}\in \mathbb{R}^d$ is randomly generated from a Gaussian noise vector, and $m_{a,t}\in\{0,1\}^d$ is randomly generated from a Bernoulli distribution for missing entries, which follows the same framework as in \cite{loh2011high}. The missing data can be said to be missing completely at random (MCAR). To handle the noisy features, we first define an oracle policy from a Bayesian perspective given the observed noisy features. Taking insights from this oracle policy, we propose an algorithm that can achieve regret bound $\tilde{O}(d\sqrt{T})$ with respect to the feature dimension $d$ and horizon time $T$. 
  
\noindent{\bf Related work.}
 \citet{auer2002using} first analyzed the linear payoff model in the bandit problem. Algorithms {\tt LinRel} \citep{auer2002using} and {\tt LinUCB} \citep{chu2011contextual} compute the expected rewards and the corresponding confidence intervals to control the exploration-exploitation trade-off and achieve an $\tilde{O}(\sqrt{dT})$ regret bound. \citet{abbasi2011improved} considered the linear bandits allowed to have infinitely many arms and proposed $\texttt{OFUL}$, which has an $\tilde{O}(d\sqrt{T})$ regret bound. However, the previous studies assumed that feature vectors are noiseless, so they cannot be directly applied to our noisy settings.
  
In our noisy settings, there exists a gap between true mean reward $z_{a,t}^\top\theta^\star$ and contaminated mean reward $x_{a,t}^\top\theta^\star$. 
We discuss some variants of contextual linear bandits for contaminated reward functions. The semi-parametric contextual bandits, where the mean reward for an arm is modeled as a linear function with a bounded confounding term that is equal for all arms, were studied by \citet{greenewald2017action,krishnamurthy2018semiparametric}, and \citet{kim2019contextual}. The algorithm proposed by \citet{krishnamurthy2018semiparametric} achieved $\tilde{O}(d\sqrt{T})$ regret in this setting. 
Another variant is the misspecified setting, where the mean rewards are allowed to have at most $\epsilon\ge 0$ distance from the best-fit linear model. In this setting, \citet{lattimore2020learning} achieved an $\tilde{O}(\sqrt{dT}+\epsilon T \sqrt{d})$ regret bound. However, proposed methods for the semi-parametric or misspecified contextual bandits show trivial regret bounds in our setting. This is because the contaminated terms in our setting differ for each arm at each time, and they are unbounded stochastic values.

Some recent studies have considered noise in feature vectors for contextual linear bandits. \citet{lamprier2018profile} considered bounded zero-mean feature noise under the assumption that true features for each arm, which are not given to the agent, are fixed over time; they achieved an $\tilde{O}(d\sqrt{T})$ regret bound. \citet{kirschner2019stochastic} did not fix the true features and considered different noise settings such that the distributions of the contexts were given to the agent each time, but the sampled contexts were hidden. They achieved an $\tilde{O}(d\sqrt{T})$ regret bound. 
However, we consider that true features are randomly sampled at each time without providing the true feature distribution to the agent, and only randomly sampled noisy features are observed. The noisy features are even allowed to have missing entries, making the problem more challenging. 
  
Several previous studies have considered how to intentionally add noise to contextual linear bandits. Differential private bandit learning \citep{shariff2018differentially,zheng2020locally} added noise to protect privacy. The authors considered adding some noise to the matrices and vectors that contain feature information. Noise vectors are generated from a Gaussian distribution with zero mean and the identity covariance matrix. Importantly, the agent knows the noise distribution to compute an upper confidence bound, which is the main difference from our setting. The algorithms suggested in \citet{shariff2018differentially} and \citet{zheng2020locally} with knowledge of the noise distribution parameters, achieved $\tilde{O}((d+d^{3/4}\epsilon^{-1/2})\sqrt{T})$ for $(\epsilon,\delta)$-differential privacy and $\tilde{O}(T^{3/4}/\epsilon)$ for $(\epsilon,\delta)$-locally differential privacy, respectively. In our noisy feature setting, we consider more complex noise generated from a multivariate Gaussian distribution that allows non-identity covariance matrices with missing entries. Moreover, we consider that the parameters of the noise distribution are not known a priori. Adversarial attacks for linear bandits were also studied by \citet{garcelon2020adversarial}, in which an adversary attempts to confuse the agent by adding noise to rewards or features. The authors suggested a way of attacking feature vectors to prevent any algorithm from achieving a sub-linear regret bound, which is the opposite of our purpose.

This paper focuses on i.i.d. feature noise randomly generated from a multivariate Gaussian distribution with missing entries for each arm. This noise cannot be handled by previous approaches, such as the semi-parametric, misspecified, or differential private bandit models, as discussed above. Interestingly, this is the first work to handle missing entries in features for contextual linear bandits as far as we know, even though missing data has been commonly studied for various learning problems such as regression for prediction and PCA \citep{shang2014robust,ramoni2001robust,han2014multiply,bang2005doubly}.

  \noindent{\bf Notation.}
 For any $A\in \mathbb{R}^{m\times m}$, we denote by $A^{-1}$ the inverse of $A$ when $A$ is invertible; otherwise, denote by $A^{-1}$ the Moore-Penrose inverse. The $i$-th singular value of $A$ is denoted by $\sigma_i (A)$. For any non-negative integer $m$, denote by $I_{m}$ the $m\times m$ identity matrix and by ${\bm 0_{ m\times 1}}$ the $m$-dimensional vector with all one entries.
 For any sets $B,C \subset \{1,...,m\}$, we denote by $A_{B,C}$ the submatrix of $A$ with row and column indexes in $B,C$, respectively. Then $A_{B,B}^{-1}$ denotes the inverse matrix of $A_{B,B}$. For any $x\in\mathbb{R}^m$, $x_B$ is a subvector of $x$ with indexes in $B$. For any $x\in\mathbb{R}^m$ and $y\in\mathbb{R}^n$, denote by $[x;y]\in\mathbb{R}^{m+n}$ the concatenated vector. Lastly, we define $\|x\|_A=\sqrt{x^\top Ax}$.

\section{PROBLEM STATEMENT}\label{sec:prob}
Here we describe the contextual linear bandit models with noisy features. Let $\mathcal{A}$ be a set of arms with $|\mathcal{A}|=K$. At each time $t\in[T]$, for all $a\in\mathcal{A}$, the true feature vector $z_{a,t}$ is assumed to be i.i.d. and generated to follow the Gaussian distribution $\mathcal{N}(\nu_f,\Sigma_f)$ where $\nu_f\in\mathbb{R}^d$ and $\Sigma_f\in\mathbb{R}^{d\times d}$. The noise vector $\varepsilon_{a,t}$ is also assumed to be i.i.d. and follows the Gaussian distribution $\mathcal{N}({\bm 0_{ d\times 1}},\Sigma_n)$ where $\Sigma_n\in\mathbb{R}^{d \times d}$. The noise feature vector is then defined as $x_{a,t}=(z_{a,t}+\varepsilon_{a,t})\circ m_{a,t}$ where $\circ$ is the element-wise product, $m_{a,t}$ is a masking vector such that $m_{a,t}\in\{0,1\}^d$, and each entry in $m_{a,t}$ follows a Bernoulli distribution with parameter $p$. We define an active set of arms $\A_t\subseteq \mathcal{A}$ for each time $t$ by removing some outliers in $\mathcal{A}$. 
We will describe the active set in more detail soon. Then, at each time, an agent observes $\mathcal{A}$ with noisy features $x_{a,t}$ for $a\in\mathcal{A}$ and selects an arm $a_t\in\A_t$ and receives reward feedback $y_t=z_{a_t,t}^\top \theta^\star+\eta_t$ with a latent parameter $\theta^\star\in \mathbb{R}^d$ where noise $\eta_t$ is i.i.d. and follows a $\sigma$- sub-Gaussian distribution with mean zero. We note that the agent has no a priori knowledge of the noise distribution parameters $\nu_f, \Sigma_f, \Sigma_n,$ and $p$, true feature vectors $z_{a,t}$, and latent model parameter $\theta^\star$. The agent can observe noisy features $x_{a,t}$ for all $a\in\mathcal{A}$. For simplicity, we assume that $\|\nu_f\|_2\le 1$, $\|\theta^\star\|\le 1$, and $\sigma=1$. In addition, we assume that $\sigma_1(\Sigma_f+\Sigma_n)$ and $\sigma_d(\Sigma_f+\Sigma_n)$ are strictly positive constants.


The active action set $\A_t$ is a subset of $\mathcal{A}$ in which all outliers are removed based on observed features such that  
\begin{align*}
\A_t=\left\{a\in\mathcal{A}; \|x_{a,t}\|_2=O\left(\sqrt{\|m_{a,t}\|_2^2\log(KT)}\right)\right\}.    
\end{align*}
 Since the features follow a Gaussian distribution, the mean reward for an arm may become large, which can induce large regret. Therefore, we devise the active action set $\A_t$ to restrict the available regret scale for our theoretical analysis, and show that $\A_t$ is the same as $\mathcal{A}$ for all $t$ with high probability, to justify the scaling condition of $\mathcal{A}_t$ .
Since every $x_{a,t}$ is an i.i.d. Gaussian random variable, from Theorem 1 in \citet{hsu2012tail} we can show that for all $a\in\mathcal{A}$ and $t\in[T]$,  given $\|m_{a,t}\|_2^2$, with probability at least $1-1/T$, we have 
\begin{align*}
\|x_{a,t}\|_2 =O\left(\sqrt{\|m_{a,t}\|_2^2\log(KT)}\right),    
\end{align*}
 which implies that $\A_t$ equals $\mathcal{A}$ for all $t\in[T]$ with high probability.

\noindent{\bf Objective function.} The goal of this problem is to design a policy that minimizes regret over a time horizon, which is defined as the difference between the cumulative reward from the optimal policy and the suggested policy. 
Therefore, it is essential to find the optimal arm at each time. In this problem, we consider a Bayesian perspective for the oracle defined on the observed information, which can be a general framework. For example, in the standard linear bandit where the true features $z_{a,t}$ are observed such that $x_{a,t}=z_{a,t}$, the standard oracle can be interpreted from a Bayesian perspective so that $a_t^\star\in \arg\max_{a\in\mathcal{A}_t}\mathbb{E}[z_{a,t}^\top\theta^\star|x_{b,t}; b\in\mathcal{A}]=\arg\max_{a\in\mathcal{A}_t}\mathbb{E}[z_{a,t}^\top\theta^\star|z_{b,t}; b\in\mathcal{A}]=\arg\max_{a\in\mathcal{A}_t} z_{a,t}^\top\theta^\star$. In our noisy setting, the agent can only observe noisy features for each arm. Therefore, determining the Bayesian oracle from noisy observed features is non-trivial. We assume that the oracle has complete knowledge of the latent parameters $\nu_f,\Sigma_f,\Sigma_n,$ and $\theta^\star$. Then, given $x_{a,t}$ for $a\in \mathcal{A}$, the oracle's action at time $t$ from a Bayesian point of view is defined as: 
\begin{align*}
	a^\star_t &\in \arg \max_{a\in  \A_t} \mathbb{E} \left[ z_{a,t}^\top \theta^\star  |  x_{b,t}; b\in\mathcal{A} \right]\cr &=\arg \max_{a\in  \A_t} \mathbb{E} \left[ z_{a,t}^\top \theta^\star  |  x_{a,t} \right],
\end{align*}
where the equality comes from the independence of the observed features.  We note that $\argmax_{a\in \mathcal{A}_t}z_{a,t}^\top\theta^\star$ can be too strong for an oracle in our setting, which is not achievable in general to obtain non-trivial regret.  This is because $z_{a,t}$ cannot be perfectly recovered from $x_{a,t}$ in general due to unknown independent noise $\varepsilon_{a,t}$ and missing entries for $a\in\mathcal{A}_t$ at each time $t$. 

Adopting the Bayesian optimal arm $a_t^\star$, we define regret for a policy of selecting arm $a_t$ at time step $t\in[T]$ as follows:
\begin{align*}
R(T)= \sum_{t=1}^T \mathbb{E}\left[z_{a^\star_t,t}^\top \theta^\star-z_{a_t,t}^\top \theta^\star\right]. 
\end{align*}

\noindent{\bf Challenges.}
Now, we explain why it is non-trivial to find the Bayesian oracle in our setting even without missing entries where $x_{a,t}=z_{a,t}+\varepsilon_{a,t}$. The true feature is $z_{a,t}=x_{a,t}-\varepsilon_{a,t}$ so the oracle policy is $a^\star_t\in\arg \max_{a\in  \A_t} \mathbb{E} \left[ z_{a,t}^\top \theta^\star  |  x_{a,t} \right]=\arg\max_{a\in\mathcal{A}_t}\{x_{a,t}^\top\theta^\star -\mathbb{E}[\varepsilon_{a,t}^\top\theta^\star|x_{a,t}]\}.$ Since $\varepsilon_{a,t}$ and $x_{a,t}$ are \textit{not} independent, we can observe $\mathbb{E}[\varepsilon_{a,t}^\top\theta^\star|x_{a,t}]\neq0$ in general, which means that the optimal arm is highly dependent on the distribution of $\varepsilon_{a,t}$ and $z_{a,t}$. Therefore, the Bayesian oracle $a_t^\star\neq \argmax_{a\in\mathcal{A}_t}x_{a,t}^\top \theta^\star$ in general, which makes our problem different from standard linear bandits where $x_{a,t}=z_{a,t}$.

\begin{figure}[ht]
    \centering
    \includegraphics[width=10cm]{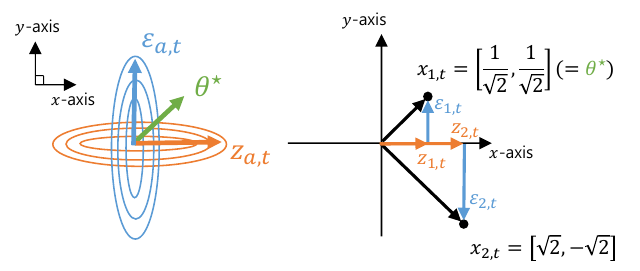} %
\caption{A toy example}    \label{fig:toy}%
\end{figure}
To illustrate this, we provide a toy example in Figure~\ref{fig:toy}. We consider Gaussian distributions for $z_{a,t}$ and $\varepsilon_{a,t}$ in two-dimensional space such that $z_{a,t}$ follows a Gaussian distribution with $v_f=[0,0]$ and $\Sigma_f=[1,0;0,0]$ and $\varepsilon_{a,t}$  follows a Gaussian distribution with mean $[0,0]$ and $\Sigma_n=[0,0;0,1]$. We note that $z_{a,t}$ and $\varepsilon_{a,t}$ are independent and orthogonal to each other because $z_{a,t}$ is on the $x$-axis and $\varepsilon_{a,t}$ is on the $y$-axis with probability $1$. We assume that there are two arms, $\{1,2\}$, and the observed features are $x_{1,t}=[1/\sqrt{2},1/\sqrt{2}]$ and $x_{2,t}=[\sqrt{2},-\sqrt{2}]$, respectively. We consider $\theta^\star=[1/\sqrt{2},1/\sqrt{2}]$, which points in the same direction as $x_{1,t}$. Then in the standard linear bandits where the observed feature is the true feature ($x_{a,t}=z_{a,t}$), arm $1$ is the optimal arm from the fact that $\mathbb{E}[z_{1,t}^\top\theta^\star|x_{1,t}]=x_{1,t}^\top \theta^\star=1$ and $\mathbb{E}[z_{2,t}^\top\theta^\star|x_{2,t}]=x_{2,t}^\top \theta^\star=0$. However, this is not the same in our noisy setting where $x_{a,t}=z_{a,t}+\varepsilon_{a,t}$. Since $z_{a,t}$ is on the $x$-axis and $\varepsilon_{a,t}$ is on the $y$-axis, we can estimate $z_{a,t}$ by projecting $x_{a,t}$ on the $x$-axis. Therefore, we obtain $\mathbb{E}[z_{1,t}|x_{1,t}]=[1/\sqrt{2},0]$ and $\mathbb{E}[z_{2,t}|x_{2,t}]=[\sqrt{2},0]$. Then the Bayesian optimal arm is arm $2$ because $\mathbb{E}[z_{1,t}^\top\theta^\star|x_{1,t}]=1/2$ and $\mathbb{E}[z_{2,t}^\top\theta^\star|x_{2,t}]=1$. From this example, we can observe that our noisy problem can be totally different from the standard linear bandits where the optimal action is directly determined by observed features. In addition, considering missing entries increases the difficulty of the problem. 

As shown in the toy example, we can observe that the optimal arm is highly dependent not only on $x_{a,t}$ and $\theta^\star$, but also on the distribution of $z_{a,t}$ and $\varepsilon_{a,t}$. Therefore, the first difficulty is deriving the Bayesian oracle policy. Then, the second difficulty is learning this oracle in a bandit setting where the latent parameters are unknown.

\noindent{\bf Contribution.}
First, we analyze the simple case with $p=1$ where there are no missing entries in the observed noisy features. Analyzing the Bayesian oracle strategy, we show that slightly modified \texttt{OFUL} \citep{abbasi2011improved} can achieve an $\tilde{O}(d\sqrt{T})$ regret bound. Then we dive into the more complicated case with missing entries. We show that the Bayesian oracle strategy requires solving non-linear programming in this case. Based on the insight, we design an algorithm that can achieve $\tilde{O}(d\sqrt{T})$ when $K$ is large. Lastly, we examine our algorithm using synthetic and real-world datasets and observe that the results are consistent with our theoretical analysis. 

\section{WARM-UP: NOISY FEATURES WITHOUT MISSING ENTRIES}\label{sec:gaussian}
In this section, we analyze the simple case of $p=1$ where the observed features have no missing entries but only Gaussian noise. First, we analyze the oracle strategy and then provide an algorithm with regret analysis.

\noindent{\bf Bayesian oracle strategy.} We examine the strategy of a Bayesian oracle that has complete knowledge of latent parameters $\nu_f$, $\Sigma_{f}$, $\Sigma_{n}$, and $\theta^\star$ in the following proposition.
\begin{proposition}
\label{thm:noisyG} 
 Define:
\begin{align*}
    \theta^\prime:= (\Sigma_{f} + \Sigma_{n})^{-1}\Sigma_{f} \theta^\star \text{ and } \overline{\theta}:=[\nu_f^\top\theta^\star-\nu_f^\top\theta^\prime;\theta^\prime].
\end{align*}
 Then the Bayesian oracle policy under the noisy features without missing entries selects arm $a^\star_t$ at time step $t\in[T]$ such that:
\begin{align*}
	a^\star_t \in \arg \max_{a\in \A_t} \mathbb{E} [ z_{a,t}^\top \theta^\star  |  x_{a,t}]  &=\arg \max_{a\in \A_t}[1;x_{a,t}]^\top \overline{\theta}\cr &= \arg \max_{a\in \A_t} x_{a,t}^\top \theta^\prime. 
\end{align*}
\end{proposition}
\begin{proof}
At each time step $t$, given noisy feature vectors, the Bayesian optimal decision is expressed as: 
\begin{align*}
    a^\star_t &\in  \arg \max_{a\in \A_t} \mathbb{E}[z_{a,t} ^\top \theta^\star| x_{b,t}; \forall b\in \mathcal{A}] \cr & = \arg \max_{a\in \A_t} \mathbb{E}[z_{a,t} | x_{a,t}]^\top \theta^\star.
\end{align*}
 From the Bayesian analysis of multivariate Gaussian random variables (Section 3.1 in \citet{smith2006optimizer}), we can show that for any $a\in\A_t$,
\begin{align}
    \mathbb{E}[z_{a,t} | x_{a,t}]^\top \theta^\star 
    &=  (x_{a,t} -\nu_f)^\top (\Sigma_{f} + \Sigma_{n})^{-1}\Sigma_{f} \theta^\star + \nu_f^\top \theta^\star \cr &=x_{a,t}^\top\theta^\prime-\nu_f^\top\theta^\prime+\nu_f^\top \theta^\star=[1;x_{a,t}]^\top \overline{\theta},
    \label{eq:bayere}
\end{align}
which concludes the proof.
\end{proof}

From Proposition~\ref{thm:noisyG}, we can observe that $\mathbb{E}[\varepsilon_{a,t}^\top \theta^\star|x_{a,t}]=x_{a,t}^\top\theta^\star-\mathbb{E}[z_{a,t}^\top \theta^\star|x_{a,t}]=x_{a,t}^\top\theta^\star-[1;x_{a,t}]^\top \overline{\theta}\neq0$ in general, as we discussed in the previous section, which makes the problem different from the standard contextual linear bandits. Although $\mathbb{E}[\varepsilon_{a,t}^\top \theta^\star|x_{a,t}]\neq0$, we can observe that interestingly, the noisy contextual bandit without missing entries still models the linear reward with respect to the modified contextual vector $[1;x_{a,t}]$, because the mean reward of an arm $a$ given the observed feature is
    $\mathbb{E}[z_{a,t}^\top \theta^\star|x_{a,t}]=[1;x_{a,t}]^\top \overline{\theta}$ from $\eqref{eq:bayere}$.
Thus, the regret of an algorithm selecting arm $a_t$ at time step $t=1,\dots,T$ can be expressed as:
\begin{align*}
R(T) &=\sum_{t=1}^T \mathbb{E}\left[z_{a^\star_t,t}^\top \theta^\star-z_{a_t,t}^\top \theta^\star\right] \cr & = \sum_{t=1}^T \mathbb{E}\left[[1;x_{a^\star_t,t}]^\top \overline{\theta} - [1;x_{a_t,t}]^\top \overline{\theta}\right].
\end{align*}

\noindent{\bf Algorithm and regret analysis.} 
From \eqref{eq:bayere}, the problem can be converted to estimating $\overline{\theta}$ using observed features $[1;x_{a,t}]$ and reward. The observed reward can be considered as \textit{Bayesian reward} whose mean reward is $[1;x_{a_t,t}]^\top \overline{\theta}$ given $x_{a,t}$'s. From the insight of the Bayesian analysis, we suggest using the strategy \texttt{OFUL} \citep{abbasi2011improved} that was proposed for standard contextual linear bandits. The algorithm is based on the principle of optimism in the face of uncertainty for reward estimation. 
To build the confidence set to estimate $\overline{\theta}$ in our setting,
we consider $\|x_{a,t}\|_2=O(\sqrt{d\log(KT)})$ for $a\in\A_t$ and $\|\overline{\theta}\|_2\le C_2$ for some constant $C_2>0$ and tune the bias term $\lambda>0$ to control the variance of the feature vector. Define the confidence set as follows: 
\begin{align*}
\mathcal{C}_t =&\{\theta\in\mathbb{R}^{d+1}: \cr & \: \|\widehat{\theta}_t-\theta\|_{V_t}\le C_1\sqrt{(d+1)\log((1+t)T)}+C_2\lambda^{1/2}\},
\end{align*}
where bias term $\lambda=C_3d\log(KT)$, $V_t=\lambda I_{d+1}+\sum_{s=1}^t [1;x_{a_t,t}][1;x_{a_t,t}]^\top$, and $\widehat{\theta}_t=V_t^{-1}\sum_{s=1}^t[1;x_{a_t,t}] y_t$ for some constants $C_1, C_3 >0$. We provide a pseudocode of the algorithm in Appendix~\ref{app:code_oful}.

 The following proposition shows that \texttt{OFUL} can achieve a tight regret bound in the noisy setting without missing entries.
\begin{proposition}\label{cor:LinRel}
{\tt OFUL} with $\mathcal{C}_{t-1}$ and modified feature vectors $[1;x_{a,t}]$  for all $a\in \mathcal{A}$ and $t\in[T]$ can guarantee $R(T)= \tilde{O}(d\sqrt{T})$.
\end{proposition}
Following the proof steps of Theorem 3 in \citet{abbasi2011improved} easily reveals the regret bound. We prove Proposition~\ref{cor:LinRel} in Appendix~\ref{app:prop_regret}.

\noindent{\bf Tightness of the regret bound.} The result in Proposition~\ref{cor:LinRel}, $\tilde{O}(d\sqrt{T})$, matches the regret lower bound for standard linear bandits, $\Omega(d\sqrt{T})$ \citep{li2010contextual}, up to poly-logarithmic factors, although our regret is defined based on the Bayesian oracle, unlike previous bandit literature. For the regret lower bound of the standard model, we additionally consider that the norms of true features are bounded by $\tilde{O}(\sqrt{d})$ as in our setting, which scales the regret $\sqrt{d}$ times more compared to the original form in \citet{li2010contextual}. Furthermore, our regret bound is comparable to the results of previous work such as the achieved $\tilde{O}(d\sqrt{T})$ regret bound considering the contaminated mean reward \citep{krishnamurthy2018semiparametric}, or $\tilde{O}(d\sqrt{T})$ \citep{lamprier2018profile,kirschner2019stochastic,shariff2018differentially} regret bounds considering feature noise. However, it is an open problem to obtain the problem-specific regret lower bound for our noisy setting under the Bayesian perspective.

\section{NOISY FEATURES WITH MISSING ENTRIES}\label{sec:gausam}

Now, we move on to our main contribution. Here, we consider a case where every component of each observed feature vector has been erased with \textit{missing probability} $1-p \in (0,1)$. Each component of the feature vector is masked with a random variable following a Bernoulli distribution with parameter $p$. Similar to the previous section, we first derive the Bayesian oracle strategy and suggest an algorithm with regret analysis.

\noindent{\bf Bayesian oracle strategy.}  Missing entries affect the information available to the oracle. The Bayesian oracle maximizes the expected reward, knowing the observed feature vectors with missing entries.
Given a feature vector $x \in \RR^d$, we define ${\cal S}(x)$ as the indexes for non-missing entries and ${\cal U}(x)$ as the indexes for missing entries. For simplicity, we use $\mathcal{S}$ for $\mathcal{S}(x)$ and $\mathcal{U}$ for $\mathcal{U}(x)$ if there is no confusion. 
Then, the Bayesian oracle strategy arises from the following theorem. 

\begin{theorem}\label{thm:noisyGerased}For any $x \in \RR^d, \nu_f\in \RR^d,$ and $\Sigma \in \RR^{d \times d}$, define $\overline{x}(\nu,\Sigma,x)\in\mathbb{R}^d$ as:
\begin{align*}
    \overline{x}(\nu,\Sigma,x)_{\mathcal{S}}&:=x_{\mathcal{S}}  \quad\mbox{and} \cr   \overline{x}(\nu,\Sigma,x)_{\mathcal{U}}&:= \nu_{\mathcal{U}} +
\Sigma_{\mathcal{U},\mathcal{S}}\Sigma_{\mathcal{S},\mathcal{S}}^{-1} (x -\nu)_{\mathcal{S}}.
\end{align*}
Then the Bayesian oracle strategy under the noisy features with missing entries selects arm $a^\star_t$ at time step $t\in[T]$ such that:
\begin{align}
	a^\star_t &\in \argmax_{a\in \A_t} \mathbb{E} [ z_{a,t}^\top \theta^\star  |  x_{a,t} ] \nonumber \cr &=\argmax_{a\in\A_t} [1;\overline{x}(\nu_f, \Sigma_f + \Sigma_n,x_{a,t})]^\top\overline{\theta}\cr & =  \argmax_{a\in \A_t} \overline{x}(\nu_f,\Sigma_{f}+\Sigma_{n},x_{a,t})^\top \theta^\prime.
\end{align}
\end{theorem}
We note that when $S=[d]$, we observe $\overline{x}(\nu_f,\Sigma_f+\Sigma_n,x_{a,t})=x_{a,t}$, which is the same as in the case of noisy features without missing entries. When $S=\emptyset$, we observe $\overline{x}(\nu_f,\Sigma_f+\Sigma_n,x_{a,t})=\nu_f$, which implies that the oracle estimates true features from the mean value directly because the oracle does not have noisy feature information.   
\begin{proof}
  Denote by $x^\prime_{a,t}$ the noisy feature vector for arm $a$ without masking such that $x^\prime_{a,t}=z_{a,t}+\varepsilon_{a,t}
$. Then we observe that $x_{a,t}^\prime$ follows the Gaussian distribution with mean $\nu_f$ and covariance matrix $\Sigma_{f}+\Sigma_{n}$. 
Then by using the conditional Gaussian distribution from Proposition 3.13 in \citet{eaton1983multivariate}, we can show that for all $a\in\A_t$, 
\begin{align}
    \mathbb{E}[x^\prime_{a,t} | x_{a,t} ] = \overline{x}(\nu_f, \Sigma_f + \Sigma_n,x_{a,t}),\label{eq:x_est}
\end{align}
which is the estimated feature vector for recovering missing values.
 Thus, from~\eqref{eq:bayere}, \eqref{eq:x_est}, and $\overline{\theta}=[\nu^\top\theta^\star-\nu^\top\theta^\prime;\theta^\prime]$, the expected reward given the observed feature vector can be expressed as
\begin{align}
\mathbb{E}[z_{a,t}^\top\theta^{\star}|x_{a,t}]
&=\mathbb{E}\left[\mathbb{E}\left[z_{a,t}^\top\theta^{\star}|x^\prime_{a,t}\right]|x_{a,t}\right] \cr &=\mathbb{E}\left[(x^\prime_{a,t}-\nu_f)^\top\theta^\prime+\nu_f^\top\theta^{\star}|x_{a,t}\right] \cr &=\mathbb{E}[x^\prime_{a,t}| x_{a,t} ]^\top \theta^\prime +\nu_f^\top(\theta^{\star}-\theta^\prime) \cr
&=  \overline{x}(\nu_f, \Sigma_f + \Sigma_n,x_{a,t})^\top \theta^\prime+\nu_f^\top(\theta^{\star}-\theta^\prime)\cr &= [1;\overline{x}(\nu_f, \Sigma_f + \Sigma_n,x_{a,t})]^\top\overline{\theta},\label{eq:r_missing}
\end{align}
which concludes the proof. 
\end{proof}
From Theorem~\ref{thm:noisyGerased}, the Bayesian oracle strategy selects an arm based on $\overline{x}(\nu_f,\Sigma_f+\Sigma_n,x_{a,t})$, which implies that the model is non-linear with respect to the observed feature $x_{a,t}$. This is a significant difference from the simple case of noisy features without missing entries. 
From the oracle strategy in Theorem~\ref{thm:noisyGerased} and \eqref{eq:r_missing}, we can show that regret can be expressed as: 
\begin{align*}
R(T) &=\sum_{t=1}^T \mathbb{E}\left[z_{a^\star_t,t}^\top \theta^\star-z_{a_t,t}^\top \theta^\star\right]\cr &=\sum_{t=1}^T \mathbb{E} \left[ [1;\overline{x}(\nu_f,\Sigma_{f}+\Sigma_{n},x_{a^\star_t,t})]^\top \overline{\theta} \right.\cr &\left. \qquad\qquad - [1;\overline{x}(\nu_f,\Sigma_{f}+\Sigma_{n},x_{a_t,t})]^\top \overline{\theta} \right].
\end{align*}

\noindent{\bf Algorithms and regret analysis.} From Theorem~\ref{thm:noisyGerased}, the oracle policy is not linear to the observed feature vectors, which implies that we cannot naively use  {\tt OFUL} to guarantee a sub-linear regret bound when the distribution latent parameters are not given to the algorithm. Therefore, we propose Algorithm~\ref{alg:alg1}, which includes a procedure to estimate \textit{Bayesian features} $[1;\overline{x}(\nu_f,\Sigma_f+\Sigma_n,x_{a,t})]$.

\begin{algorithm}[t]
   \caption{Contextual Linear Bandits on Bayesian Features ({\tt CLBBF})}
    \label{alg:alg1} 
\begin{algorithmic}
\STATE {\bfseries Initialize:}  $Z \leftarrow {\bm 0}_{d\times d}$, $\xi \leftarrow {\bm 0}_{d\times 1}$, 
$n \leftarrow 0$, $i\leftarrow 1$.
\STATE $n \leftarrow n \,+$ the total number of non-missing entries in $x_{a,1}$ for all $a\in\mathcal{A}$.
\STATE 
 $Z \leftarrow Z+ \sum_{a\in\mathcal{A}} x_{a,1}x_{a,1}^\top$, $\xi \leftarrow \xi + \sum_{a\in \mathcal{A}} x_{a,1}$.
\STATE Select $a_1$ uniformly at random in $\A_1$.
\STATE Observe reward $y_1$.
\FOR{$t =2$ {\bfseries to} $T$}
\STATE  $n \leftarrow n \,+$ the total number of non-missing entries in $x_{a,t}$ for all $a\in\mathcal{A}$.
\STATE
 $Z \leftarrow Z+ \sum_{a\in\mathcal{A}} x_{a,t}x_{a,t}^\top$, $\xi \leftarrow \xi + \sum_{a\in \mathcal{A}} x_{a,t}$.\STATE Estimate parameters: 
\STATE$\widehat{p} \leftarrow \frac{\max\{1, n \}}{td K}$, \quad  $\widehat{\nu}\leftarrow \frac{1}{tK \widehat{p}}\xi$,   \STATE$\widehat{\Sigma}\leftarrow \frac{1}{t K} Z \circ \left(\frac{\widehat{p}-1}{\widehat{p}^2} I_{d\times d}+ \frac{1}{\widehat{p}^2} {\bm 1}_{d\times d} \right) - \widehat{\nu} \widehat{\nu}^\top$.
 \STATE Estimate Bayesian features: 
 \STATE $\widehat{z}_{a,t}\leftarrow[1;\overline{x}(\widehat{\nu},\widehat{\Sigma},x_{a,t})]$ for $a\in\A_t$.
 \IF {$t=2^i$}
 \STATE Update selected Bayesian features: \STATE$\widehat{z}_{a_s,s}\leftarrow[1;\overline{x}(\widehat{\nu},\widehat{\Sigma},x_{a_s,s})]$ for all $s\in[t-1]$.
 \STATE $i\leftarrow i+1$.
 \ENDIF
 \STATE Select $a_t=\arg\max_{a\in \A_t}\max_{\theta \in  \mathcal{C}_{t-1}}\langle \widehat{z}_{a,t},\theta\rangle$.
 \STATE Observe reward $y_t$.
\ENDFOR
\end{algorithmic}
\end{algorithm}

Algorithm~\ref{alg:alg1} estimates the parameters of the distribution for the feature vectors from the information of the observed features at each time. Using the estimated distribution parameters, the algorithm estimates the Bayesian feature $[1;\overline{x}(\nu_f,\Sigma_f+\Sigma_n,x_{a,t})]$ as $\widehat{z}_{a,t}$  for all $a\in\A_t$. Furthermore, the algorithm rarely updates the estimated features for previously chosen arms using the current estimated distribution parameters while reducing the computation cost. Finally, it takes a strategy based on the principle of optimism in the face of uncertainty using estimated feature vectors $\widehat{z}_{a,t}$. The algorithm selects $a_t$ and observes reward $y_t$ for each time $t$. We note that the algorithm considers the observed reward as Bayesian reward whose mean reward is $[1;\overline{x}(\nu_f, \Sigma_f + \Sigma_n,x_{a_t,t})]^\top\overline{\theta}$ from \eqref{eq:r_missing}.

The necessary notations for Algorithm~\ref{alg:alg1} are defined as follows. First we define $\lambda=C_4d\log(KT)$,
$V_t= \lambda I_{d+1}+\sum_{s=1}^t \widehat{z}_{a_s,s}\widehat{z}_{a_s,s}^\top$, 
$\widehat{\theta}_t=V_t^{-1} \sum_{s=1}^{t}\widehat{z}_{a_s,s} y_s$, and $X_t=\sum_{s=1}^t\|\widehat{z}_{a_s,s}\|_{V_t^{-1}}$  for some large constant $C_4>0$. Note that $\|\overline{\theta}\|_2\le C_2$ for some constant $C_2>0$ and that the bias term $\lambda$ is tuned to control the variance of the observed feature norms. 
In addition, note that $\widehat{p}$ is an estimator for $p$ at each time in the algorithm. Then, we define the confidence set for estimating a latent parameter $\overline{\theta}$ as
\begin{align}
    \mathcal{C}_t &=\left\{\theta\in \mathbb{R}^{d+1}:\|\widehat{\theta}_t-\theta\|_{V_t}\le C_5\sqrt{(d+1)\log((1+t)T)}\right.\cr  &\left.\qquad\qquad +C_2\lambda^{1/2}+C_6(d/\widehat{p})^{3/2}\sqrt{\frac{\log(KT)}{K}}X_t \right\},\label{eq:conf}
\end{align}
for some large constants $C_5,C_6>0$. The estimation error of Bayesian features incurs that Algorithm~\ref{alg:alg1} requires an additional confidence bound term, $C_6(d/\widehat{p})^{3/2}\sqrt{\log(KT)/K}X_t$, in \eqref{eq:conf} compared to \texttt{OFUL}. 
By solving the convex problem to find $a_t$ in Algorithm~\ref{alg:alg1}, as stated in Section 19.3 in \citet{lattimore2020bandit}, the action in each round can be simply calculated as 
\begin{align*}
a_t&=\argmax_{a\in\A_t}\left\{\widehat{z}_{a,t}^\top \widehat{\theta}_{t-1}+\left(C_5\sqrt{(d+1)\log(tT)}+C_2\lambda^{1/2}\right.\right.\cr &\left.\left.+C_6(d/\widehat{p})^{3/2}\sqrt{\frac{\log(KT)}{K}}X_{t-1}\right)\sqrt{\widehat{z}_{a,t}^\top V_{t-1}^{-1}\widehat{z}_{a,t}}\right\}.    
\end{align*}
 Then, the following theorem provides regret for the algorithm.
\begin{theorem}
 Algorithm~\ref{alg:alg1} achieves the regret bound of
\begin{align*}
R(T)=\tilde{O}\left(d\sqrt{T}+\frac{d^2}{p^{3/2}}\sqrt{\frac{T}{K}}+\frac{d}{p^4K}\right).
\end{align*}\label{thm:regret}
\end{theorem}

 The first term for the regret bound in Theorem~\ref{thm:regret} comes from the analysis of optimism in the face of uncertainty, and the second and third terms come from the estimation error for Bayesian features, which is the main difference from standard linear bandits where the true feature vectors are given. We can note that the second term dominates over the third one when $T$ is large. An additional factor $d$ compared to Proposition~\ref{cor:LinRel} comes from the estimation error of Bayesian features, and factor $1/p^{3/2}$ mainly comes from estimation error of $\widehat{\Sigma}$.
 From Theorem~\ref{thm:regret}, when $K=\Omega(\max\{d^2/p^3,1/(\sqrt{T}p^4)\})$ 
 we get $$R(T)=\tilde{O}(d\sqrt{T}).$$ Therefore, the regret becomes insensitive to $p$ when the algorithm has sufficient observed feature information to estimate Bayesian features accurately. We provide a further discussion about the tightness of the bound with respect to $p$ in Appendix~\ref{app:p_tight}.  We also note that by using the well-known doubling trick, we can run our algorithm without information of $T$ a priori.
 

\begin{proof}[Proof sketch] We provide a proof sketch here; Appendix~\ref{app:proof_regret} contains the full version. Let $\overline{z}_{a,t}=[1;\overline{x}(\nu,\Sigma_f+\Sigma_n,x_{a,t})]$ for simplicity, and instantaneous regret $r_t= \overline{z}_{a^\star_t,t}^\top\overline{\theta}- \overline{z}_{a_t,t}^\top\overline{\theta}$. For $s\le t$, we write  $\widehat{\nu}_t$,  $\widehat{\Sigma}_t$, $\widehat{z}_{a,s}(t)$, and $\widehat{p}_t$ for $\widehat{\nu}$, $\widehat{\Sigma}$, $\widehat{z}_{a,s}$, and $\widehat{p}$ at time step $t$, respectively. From the condition for the active set, we have $\|\overline{z}_{a,t}\|_2=O(\sqrt{d\log(KT)})$ for all $a\in\A_t$. The confidence set $\mathcal{C}_t$ is designed on this constraint for $\overline{z}_{a,t}$. However, the algorithm selects an arm based on estimated features $\widehat{z}_{a,t}(t)$ rather than $\overline{z}_{a,t}$. Hence, the first few steps require collecting feature information to estimate $\overline{z}_{a_s,s}$ as $\widehat{z}_{a_s,s}(t)$ to satisfy   $\|\widehat{z}_{a_s,s}(t)\|_2=O(\sqrt{d\log(KT)})$. Let $\tau=2\lceil d(\log(T))^2/(Kp^4)\rceil$. Then, we show that for all $t>\tau$ and $1\le s\le t$, with high probability we have 
\begin{align}
\|\widehat{z}_{a_s,s}(t)\|_2=O\left(\sqrt{d\log(KT)}\right).\label{eq:x_hat_bd}    
\end{align}
 To obtain a regret bound before $\tau$, we use the fact that $x_a$ follows a Gaussian distribution given $m_{a,t}$. Using
a bound for an expected maximum Gaussian variable, we can show that 
\begin{align}
\mathbb{E} \left[ \overline{z}_{a_t^\star,t}^\top 
  \overline{\theta} - \overline{z}_{a_t,t}^\top
  \overline{\theta}\right]=O\left(\sqrt{\log (K)} \right).\label{eq:r_gauss_bd}
\end{align}
Then from \eqref{eq:r_gauss_bd}, the regret bound for the first $\tau$ time steps can be obtained as
\begin{align}
\sum_{t=1}^{\tau}\mathbb{E} \left[r_t\right]=O\left(\tau\sqrt{\log(K)}\right)=\tilde{O}\left(d/(Kp^4)\right).\label{eq:explore_regret}    
\end{align}
Let $\tilde{\theta}_t=\arg\max_{\theta\in \mathcal{C}_{t-1}}\max_{a\in \A_t}\langle \widehat{z}_{a,t}(t),\theta \rangle$. Now, we analyze regret after the time step $\tau$. Our algorithm selects an arm $a_t$ based on the estimated features $\widehat{z}_{a,t}(t)$ and observes $y_t$ whose mean reward is $\overline{z}_{a_t,t}^\top \overline{\theta}$ from the Bayesian view.  Therefore, regret is influenced by both estimation errors for $\tilde{\theta}_t$ and $\widehat{z}_{a,t}(t)$, which is the main difference from the standard linear bandits in regret analysis. First, we try to separate regret by the estimation errors; we show that with high probability,
\begin{align}
r_t&\le  \|\overline{z}_{a^\star_t,t}-\widehat{z}_{a^\star_t,t}(t)\|_2\| \overline{\theta}\|_2\ + \|\overline{z}_{a_t,t}-\widehat{z}_{a_t,t}(t)\|_2\| \overline{\theta}\|_2\cr & \qquad+\| \overline{\theta}-\tilde{\theta}_t \|_{V_{t-1}}\|\widehat{z}_{a_t,t}(t)\|_{V_{t-1}^{-1}},\label{eq:ins_reg} 
\end{align}
where the instantaneous regret bound comprises the estimation errors for $\widehat{z}_{a_t^\star,t}$, $\widehat{z}_{a_t,t}(t)$, and $\tilde{\theta}_t$.
To obtain the error bounds for $\| \overline{z}_{a,t}-\widehat{z}_{a,t}(t) \|_2$ and $\|\overline{\theta}-\tilde{\theta}_t\|_2,$ we consider the estimation errors for $\widehat{\nu}_t$ and $\widehat{\Sigma}_t$. Using vector and matrix concentration inequalities, we can show that with high probability
 \begin{align}
\|\overline{z}_{a,t} - \widehat{z}_{a,t}(t)\|_2 =O\left((d/p^{3/2})\sqrt{\log(KT)\log(T)/(tK)}\right).\label{eq:x_err}     
 \end{align}
 From \eqref{eq:x_hat_bd}, \eqref{eq:ins_reg}, \eqref{eq:x_err}, and since $\widehat{p}_t=\Theta(p)$ with high probability from Hoeffding's inequality, we can show that $\overline{\theta}\in \mathcal{C}_{t-1}$ with high probability. In addition, we show that $X_{t-1}=O\left(\sqrt{td\log T}\right).$
Then, with high probability, we have
 \begin{align}
     \|\overline{\theta}-\tilde{\theta}_t\|_{V_{t-1}}&=O\left(\sqrt{d\log(tKT)}\right.\cr &\left.+(d/p)^{3/2}\log(T)\sqrt{\log(KT)/K}\right).\label{eq:theta_err}
 \end{align}
 From \eqref{eq:ins_reg}, \eqref{eq:x_err}, and \eqref{eq:theta_err}, we have 
 \begin{align}
 r_t&=\tilde{O}\left(\left(\sqrt{d}+(d/p)^{3/2}\sqrt{1/K}\right)\|\widehat{z}_{a_t,t}(t)\|_{V_{t-1}^{-1}}\right.\cr &\left.\qquad+(d/p^{3/2})\sqrt{1/(tK)}\right).\label{eq:r_t_bd}    
 \end{align}
  Using \eqref{eq:x_hat_bd} and Lemma 11 in \cite{abbasi2011improved}, we can show that with high probability
  \begin{align}
   \sum_{t=\tau+1}^T\|\widehat{z}_{a_t,t}(t)\|^2_{V_{t-1}^{-1}}=O\left(d\log(T\log(KT))\log(T)\right).\label{eq:sum_x}   
  \end{align}
   Finally, with \eqref{eq:explore_regret}, \eqref{eq:r_t_bd}, and \eqref{eq:sum_x}, using the Cauchy-Schwarz inequality, we have
 \begin{align*}
     \sum_{t=1}^T\mathbb{E}[r_t]
     &=\tilde{O}\left(d\sqrt{T}+(d^2/p^{3/2})\sqrt{T/K}+d/(Kp^4)\right), 
 \end{align*}
 which concludes the proof.
\end{proof}


\noindent{\bf Algorithm efficiency.}  Algorithm~\ref{alg:alg1} rarely updates the estimated features for previously chosen arms, achieving the $O(T\log(T))$ computation cost over the horizon $T$. The update procedure requires $O(t)$ memory space for each time step $t$. This update is required to obtain well-estimated features for the previously chosen arms, which are then used to obtain $V_t$, $\widehat{\theta}_t$, and $X_t$.
We propose a more efficient algorithm (Algorithm~\ref{alg:alg2} in Appendix~\ref{app:eff_alg}) to reduce the computation cost and the required memory space for Algorithm~\ref{alg:alg1}, achieving the $O(T)$ computation cost and $O(1)$ memory space.
However, the algorithm requires information of $\alpha>0$ that satisfies $\alpha\le p$. We can show that this algorithm achieves $ R(T)=\tilde{O}\left(d\sqrt{T}+(d^2/p^{3/2})\sqrt{T/K}+d/(\alpha^4K)\right).$ Appendix~\ref{app:eff_alg} contains a more detailed explanation and regret analysis of the algorithm.


\section{NUMERICAL EXPERIMENTS}\label{sec:exp}

\begin{figure}[t]
\centering
\hspace{0mm}
\begin{subfigure}[b]{.44\textwidth}
\includegraphics[width=\linewidth]{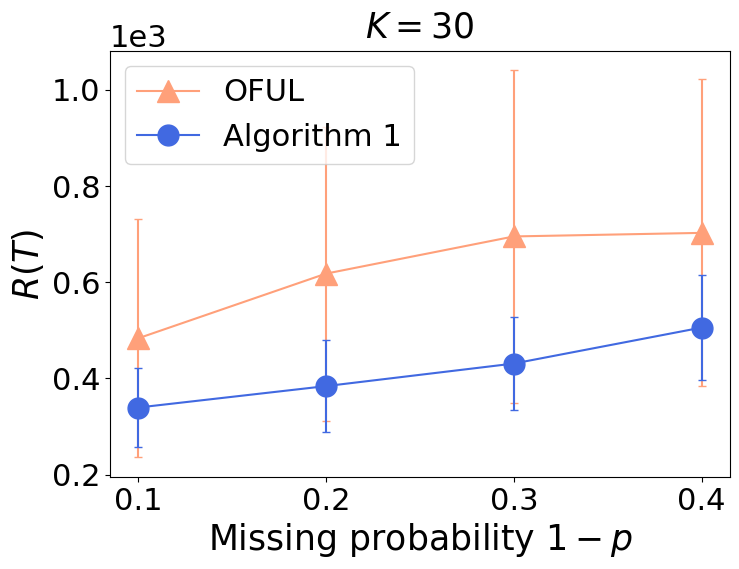}\caption{}\end{subfigure}
\begin{subfigure}[b]{.45\textwidth}\includegraphics[width=\linewidth]{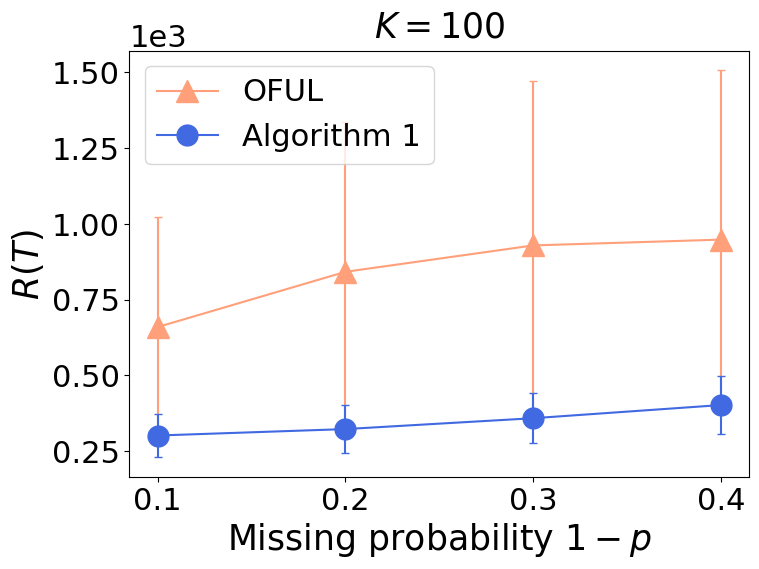}\caption{}\end{subfigure}
\begin{subfigure}[b]{.45\textwidth}\includegraphics[width=\linewidth]{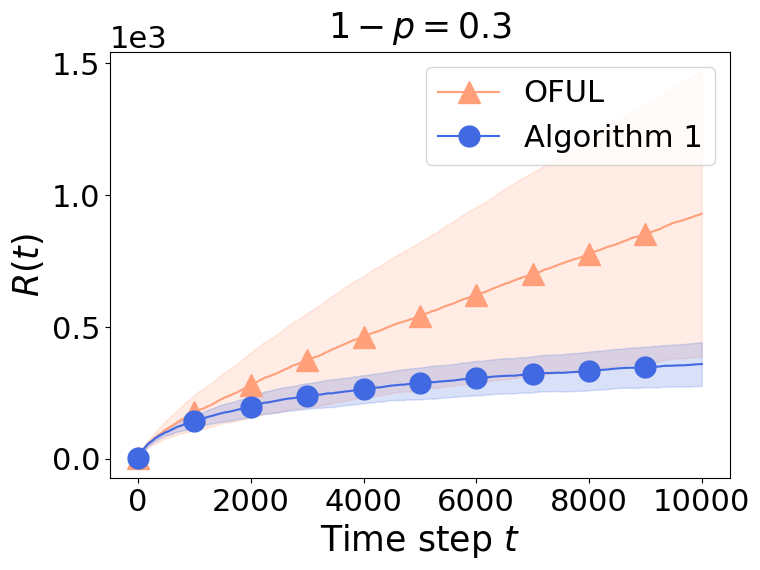}\caption{}\end{subfigure}
\hspace{-1.5mm}\begin{subfigure}[b]{.45\textwidth}
\includegraphics[width=\linewidth]{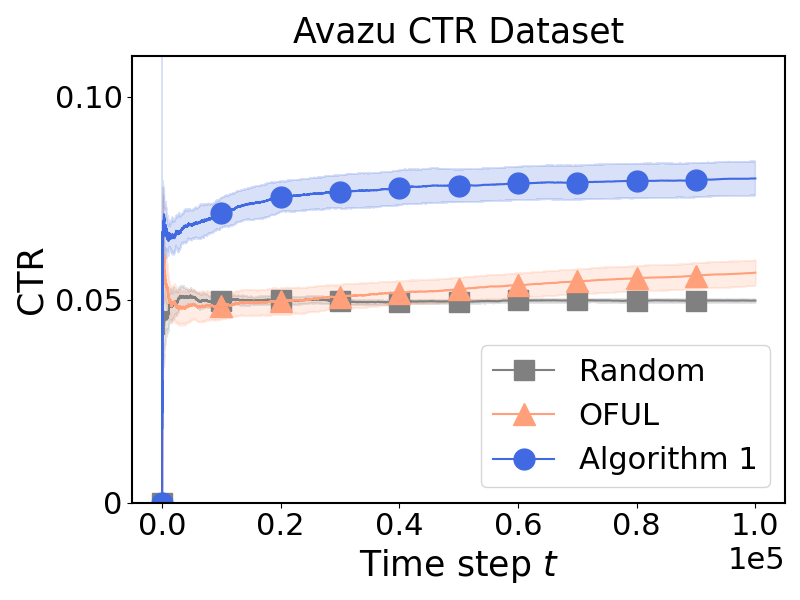}\caption{}\end{subfigure}
\caption{Performance of Algorithm~\ref{alg:alg1} and \texttt{OFUL}: $R(T)$ versus missing probability $1-p$ with
(a) $K=30$ and (b) $K=100$ in  synthetic datasets; and  $R(t)$ or CTR versus time step $t$ in  (c) synthetic and (d) real-world datasets.
}    \label{fig:syn}%
\end{figure}



In this section, we present numerical results of experiments for synthetic and real-world datasets.\footnote{Our code is available at \url{https://github.com/junghunkim7786/contextual_linear_bandits_under_noisy_features}} We repeat each experiment $10$ times. 
First, we describe the experimental setting for synthetic datasets. For the latent parameters, we generate $\theta^\star$ and $\nu_f$ from the uniform distribution in $[0,1]$ and normalize them using $2$-norm. Then, we generate random matrices $A,B\in[0,1]^{d\times d}$ from the uniform distribution and construct the covariance matrices as $\Sigma_f=A^\top A$ and $\Sigma_n=B^\top B$ with spectral norm normalization. Feature vectors are generated from $\mathcal{N}(\nu_f,\Sigma_f)$, noise vectors are generated from $\mathcal{N}(0,\Sigma_n)$, and missing entries are assigned following a Bernoulli distribution with probability $p$. Using the distributions, we generate $K$ number of noisy feature vectors for each time. The reward noise $\eta_t$  is generated from $\mathcal{N}(0,1)$. We set $d=2$ and $T=10^4$.

For real-world experiments, we use Avazu click-through rate (CTR) dataset \citep{Avazu_CTR_dataset} that contains `clicked or not' information for each advertisement recommendation. We use an autoencoder model to preprocess ad-user pair feature information for constructing item feature vectors. We set preprocessed feature dimension $d=32$, $K=20$, and $T=10^5$. We erase some entries in the preprocessed feature vectors with missing probability $1-p=0.1$. A more detailed explanation is deferred to Appendix~\ref{app:real}.

We compare the performance of Algorithm~\ref{alg:alg1}  with \texttt{OFUL} described in Section~\ref{sec:gaussian}. We first examine cumulative regret over the time horizon $T$ for each algorithm in synthetic datasets varying $1-p$ from $0$ to $0.4$, where the number of missing entries is likely to increase as the missing probability increases. Figure~\ref{fig:syn} (a,b) confirms that our algorithm performs better than \texttt{OFUL} for various $1-p$. In Figure~\ref{fig:syn} (a) when $K=30$, the regret of our algorithm increases rapidly as the missing probability increases, while in 
Figure~\ref{fig:syn} (b) when $K=100$, we can observe that our algorithm becomes more robust to the missing probability. This result is consistent with Theorem~\ref{thm:regret}, which shows that our algorithm is insensitive to $p$ with $R(T)=\tilde{O}(d\sqrt{T})$ when $K$ is sufficiently large.
Furthermore, we can observe that the regret variance for Algorithm~\ref{alg:alg1} is smaller than that for \texttt{OFUL}.

Next, in Figure~\ref{fig:syn} (c), we examine the cumulative regret over time steps $t\in[T]$ when $1-p=0.3$ and $K=100$ in synthetic datasets. We can observe that Algorithm~\ref{alg:alg1} achieves much better performance with sub-linearly increasing regret with low variance, whereas \texttt{OFUL} exhibits almost linearly increasing regret with large variance. In Figure~\ref{fig:syn} (d), we demonstrate that our algorithm achieves higher CTRs than \texttt{OFUL} in the real-world dataset. Further experiments for various real-world datasets are provided in Appendix~\ref{app:real}.

\section{DISCUSSION ON EXTENSIONS} In our setting, we consider that the features of the arms are generated from the same Gaussian distribution. 
 However, our method can be easily extended to the case where the features of the arms are generated from different Gaussian distributions with each other. In such a case, the algorithm needs to estimate latent distribution parameters for each arm. Then, with a slight modification to our algorithm, we can achieve $R(T)=\tilde{O}((d^2/p^{3/2})\sqrt{T}+d/p^4)$. 
 In addition, in our setting, we consider that the features of arms at the same time are independent. We can extend our setting to the dependent case such that the $K$ number of features is generated from a Gaussian distribution with parameters $\nu_f\in\mathbb{R}^{Kd}$ and $\Sigma_f\in \mathbb{R}^{Kd\times Kd}$ having a Gaussian noise with $\Sigma_n\in\mathbb{R}^{Kd\times Kd}$. With a slight modification of our algorithm, we can achieve $R(T)=\tilde{O}(((Kd)^2/p^{3/2})\sqrt{T}+(dK)/p^4)$. 
 
 In our setting, we consider the Gaussian conjugate distribution of features for Bayesian analysis. It would be of interest to extend our setting to other conjugate distributions.
 
\section{CONCLUSION}\label{sec:conclusion}
In this paper, we studied contextual linear bandits under noisy features that contain Gaussian noise and missing entries. We analyzed Bayesian oracles and subsequently proposed an algorithm to achieve $\tilde{O}(d\sqrt{T})$ when the number of arms is large. Lastly, we demonstrated the performance of our algorithm using synthetic and real-world datasets.

{\bf Limitation:} We leave several open questions. For instance, 
extensions to other distributions beyond the Gaussian distribution for Bayesian analysis would be of interest. In addition, obtaining a regret lower bound would be useful in understanding the fundamental limitations of our model. 

{\bf Potential negative societal impacts:}
This study focuses on theoretical analysis; therefore, we could not see any negative social consequences.

\section*{Acknowledgements}
This work was supported by Institute of Information \& communications Technology Planning \& Evaluation (IITP) grants funded by the Korea government(MSIT) (No.2022-0-00311, Development of Goal-Oriented Reinforcement Learning Techniques for Contact-Rich Robotic Manipulation of Everyday Objects; No.2019-0-00075, Artificial Intelligence Graduate School Program(KAIST)). The authors would like to  thank the anonymous reviewers for their helpful comments and suggestions.

\bibliographystyle{plainnat}
\bibliography{refs}
\newpage
\appendix


\section{APPENDIX}\label{sec:appendix}

\subsection{Pseudocode of  \texttt{OFUL}}\label{app:code_oful}

\begin{algorithm}[]
   \caption{{\tt OFUL} \citep{abbasi2011improved}}
\begin{algorithmic}
\STATE Select $a_1$ uniformly at random in $\A_1$
\STATE Observe reward $y_1$
\FOR{$t =2$ {\bfseries to} $T$}
 \STATE Select $a_t=\arg\max_{a\in \A_t}\max_{\theta\in\mathcal{C}_{t-1}} [1;x_{a,t}]^\top\theta$
\STATE Observe reward $y_t$
\ENDFOR
\end{algorithmic}
\end{algorithm}
\subsection{Proof of Proposition~\ref{cor:LinRel}}\label{app:prop_regret}
From \texttt{OFUL} in \citet{abbasi2011improved}, the algorithm selects an arm $a_t$ at each time $t$ such that
$$(a_t,\tilde{\theta}_t)=\argmax_{(a,\theta)\in \mathcal{A}_t\times \mathcal{C}_{t-1}}[1;x_{a,t}]^\top \theta.$$

For the completeness, we provide a proof following the proof steps of Theorem 3 in \citet{abbasi2011improved}. We first provide a lemma about a property of the confidence set $\mathcal{C}_{t-1}$. From Section 3.1 in \citet{smith2006optimizer}, we observe that $y_{a_t}$ given $x_{a,t}$'s follows a $C$-subGuassian distribution with mean $[1;x_{a,t}]^\top \overline{\theta}$ for some  constant $C>0$.

\begin{lemma}[Theorem 2 in \citet{abbasi2011improved}]
With probability at least $1-1/T$, for all $t\ge 0$, $\overline{\theta}$ lies in the set $\mathcal{C}_{t-1}.$\label{lem:oful_conf}
\end{lemma}
Let instantaneous regret $r_t= [1;x_{a^\star_t,t}]^\top\overline{\theta}- [1;x_{a_t,t}]^\top\overline{\theta}$. Then from Lemma~\ref{lem:oful_conf}, with probability at least $1-1/T$, for all $t\in[T]$ we have 
\begin{align}
    r_t&=[1;x_{a^\star_t,t}]^\top\overline{\theta}- [1;x_{a_t,t}]^\top\overline{\theta}\cr
    &\le [1;x_{a_t,t}]^\top \tilde{\theta}_t-[1;x_{a_t,t}]^\top\overline{\theta}\cr 
    &\le \|[1;x_{a_t,t}]\|_{V_{t-1}^{-1}}\|\tilde{\theta}_t-\overline{\theta}\|_{V_{t-1}},\label{eq:r_bd_step1}
\end{align}
where the first inequality comes from the strategy of the algorithm  for selecting $a_t$ under $\overline{\theta}\in C_{t-1}$.

From Lemma~\ref{lem:oful_conf}, we can obtain that with probability at least $1-1/T$, for all $t\in[T]$, 
\begin{align}
    \|\tilde{\theta}_t-\overline{\theta}\|_{V_{t-1}}\le C_1\sqrt{(d+1)\log(tT)}+C_2\lambda^{1/2},\label{eq:theta_gap}
\end{align}
where $\lambda=C_3d\log(KT)$ for some constants $C_1, C_2,C_3>0$. We denote by $E$ the event of \eqref{eq:theta_gap}.  From \eqref{eq:r_bd_step1}, \eqref{eq:theta_gap}, and the fact that $r_t=O(\sqrt{d\log(KT)})$ from the scale condition for $\mathcal{A}_t$, we can obtain
\begin{align}
    \sum_{t=1}^T\mathbb{E}[r_t] =\tilde{O}\left(\sqrt{d}\sum_{t=1}^T\mathbb{E}\left[\|[1;x_{a_t,t}]\|_{V_{t-1}^{-1}}\mid E\right]+\sqrt{d}\right).\label{eq:r_bd_step2}
\end{align}
In what follows, we provide a lemma for a bound of $\sum_{t=1}^T\|[1;x_{a_t,t}]\|_{V_{t-1}^{-1}}$.
\begin{lemma} We have
$$\sum_{t=1}^T\|[1;x_{a_t,t}]\|_{V_{t-1}^{-1}}^2=O\left( d\log(T\log(KT))\right)$$\label{lem:x_sum_oful}
\end{lemma}
\vfill

\begin{proof}
 From the scale condition of $\|x_{a,t}\|_2$ for $a\in\A_t$, there exists a constant $C>0$ such that
\begin{align*}
    \|[1;x_{a_t,t}]\|_2^2\le Cd\log(KT)
\end{align*}
Then for sufficiently large $C_3>0$ with $\lambda=C_3d\log(KT)$, we have
\begin{align}
    \|[1;x_{a_t,t}]\|_{V_{t-1}^{-1}}^2\le \|[1;x_{a_t,t}]\|_2^2\|V_{t-1}^{-1}\|_2\le \|[1;x_{a_t,t}]\|_2^2/\lambda=\|[1;x_{a_t,t}]\|_2^2/(C_3d\log(KT))\le 1.\label{eq:x_V_bd}
\end{align}
From \eqref{eq:x_V_bd}, by following the proof steps in Lemma 11 in \citet{abbasi2011improved}, we have
\begin{align}
&\sum_{t=1}^T\|[1;x_{a_t,t}]\|_{V_{t-1}^{-1}}^2\le 2\sum_{t=1}^T\log(1+\|[1;x_{a_t,t}]\|_{V_{t-1}^{-1}}^2)\cr &\le 2\log\det(V_T)\le 2(d+1)\log(C(T+1)\log(KT)). \label{eq:x_sum_bd_oful}
\end{align}

\end{proof}
Finally from \eqref{eq:r_bd_step2} and Lemma~\ref{lem:x_sum_oful}, we can obtain 
\begin{align*}
    R(T)&=\mathbb{E}\left[\sum_{t=1}^Tr_t\right]\cr &=\tilde{O}\left( \sqrt{d}\mathbb{E}\left[\sum_{t=1}^T\|[1;x_{a_t,t}]\|_{V_{t-1}^{-1}}\mid E\right]+\sqrt{d}\right)\cr &\le\tilde{O}\left(\sqrt{d}\mathbb{E}\left[\sqrt{T\sum_{t=1}^T\|[1;x_{a_t,t}]\|^2_{V_{t-1}^{-1}}}\mid E\right]+\sqrt{d}\right)\cr &=\tilde{O}(d\sqrt{T}).
\end{align*}
{\bfseries Further discussion.} \texttt{OFUL} achieved $\tilde{O}(d\sqrt{T})$ \cite{abbasi2011improved} in the standard linear bandits, with mean reward scale bounded by $1$. Therefore,  it may be of interest that \texttt{OFUL}, in our setting, still achieves $\tilde{O}(d\sqrt{T})$ for the case of mean reward scale $|[1;x_{a,t}]^\top\overline{\theta}|=\tilde{O}(\sqrt{d})$ with $\|[1;x_{a,t}]\|_2=\tilde{O}(\sqrt{d})$ and $\|\overline{\theta}\|_2=O(1)$. This can be obtained from that we tune a bias term $\lambda$ properly for dealing with the variance of $\|[1;x_{a_t,t}]\|_{V_{t-1}^{-1}}^2$ shown in \eqref{eq:x_V_bd}.

\subsection{Proof of Theorem~\ref{thm:regret}}\label{app:proof_regret}

For the notational simplicity, let $\Sigma = \Sigma_{f}+\Sigma_{n}$ and $\overline{z}_{a,t}=[1;\overline{x}(\nu,\Sigma,x_{a,t})]$. Also, for $s\le t$, we write  $\widehat{z}_{a,s}(t)$ for  $\widehat{z}_{a,s}$ at time step $t$. From the condition for the active set, we observe that $\|\overline{z}_{a,t}\|_2=O(\sqrt{d\log(KT)})$. The confidence set $\mathcal{C}_t$ is designed considering this constraint to get a tight bound. However, the algorithm selects an arm based on estimated features $\widehat{z}_{a,t}(t)$ instead of $\overline{z}_{a,t}$. Hence, for the first few steps, it requires to collect feature information to estimate $\widehat{z}_{a_s,s}(t)$ for satisfying   $\|\widehat{z}_{a_s,s}(t)\|_2=O(\sqrt{d\log(KT)})$. Recall that $\theta^\prime= (\Sigma_{f} + \Sigma_{n})^{-1}\Sigma_{f} \theta^\star$ and  $\overline{\theta}=[\nu^\top\theta^\star-\nu^\top\theta^\prime;\theta^\prime].$ Let $\tau=2\lceil d(\log(T))^2/Kp^4\rceil$. Then for any $t> \tau$ and $0<s\le t$ we have $\|\widehat{z}_{a_s,s}(t)\|_2=O(\sqrt{d\log(KT)})$ with a high probability, which will be shown later.
The regret of Algorithm~\ref{alg:alg1} can be decomposed into the followings:
\begin{align}
R(T)=& \sum_{t=1}^T \mathbb{E} \left[ [1;\overline{x}(\nu_f,\Sigma,x_{a^\star_t,t})]^\top \overline{\theta}- [1;\overline{x}(\nu_f,\Sigma,x_{a_t,t})]^\top \overline{\theta}\right]  \cr
=& \sum_{t=1}^{\tau} \mathbb{E} \left[ \overline{x}(\nu_f,\Sigma,x_{a^\star_t,t})^\top \theta^\prime - \overline{x}(\nu_f,\Sigma,x_{a_t,t})^\top \theta^\prime\right]\cr &\quad+ \sum_{t=\tau+1}^{T} \mathbb{E} \left[ [1,\overline{x}(\nu_f,\Sigma,x_{a^\star_t,t})]^\top \overline{\theta} - [1,\overline{x}(\nu_f,\Sigma,x_{a_t,t})]^\top \overline{\theta}\right].\cr \label{eq:overRT}
\end{align}
We use the following lemma to get a regret bound before $\tau$ time steps. 
\begin{lemma} For any time $t>0$, we have
\begin{align}
&\mathbb{E} \left[ \overline{x}(\nu_f,\Sigma,x_{a^\star_t,t})^\top 
  \theta^\prime - \overline{x}(\nu_f,\Sigma,x_{a_t,t})^\top
  \theta^\prime\right]=O\left(\sqrt{\log (K)} \right).
\end{align}\label{lem:max_r}
\end{lemma}
\begin{proof}
For any given $m_{a,t}$, 
$\overline{x}(\nu_f,\Sigma,x_{a,t})^\top\theta^\prime $ is a random variable with a Gaussian distribution which mean is $\nu_f^\top\theta^\prime$ for all $a\in \mathcal{A}$. For analyzing the random variable, we provide a bound for the expected maximum value of random variables according to an i.i.d Gaussian distribution in the following.
\begin{lemma} Let $X_i$ be an independent random variable with $\mathcal{N}(\mu,\sigma_i^2)$ for all $i\in\{1,...,n\}$. Define $X^\star=\max\limits_{i\in\{1,\dots,n\}}X_i$ and $\sigma^{\star}=\max\limits_{i\in\{1,\dots,n\}}\sigma_i$. Then, \[\mathbb{E}[X^\star]\le\mu+ \sigma^{\star} \sqrt{2\log(n)}.\]\label{lem:maxgaussian} 
\end{lemma}
\begin{proof}
 For any $\lambda\ge0$,
\begin{align*}
    \exp(\lambda\mathbb{E}[X^\star])&\le \mathbb{E}[\exp(\lambda X^\star)]\cr&=\mathbb{E}\left[\max_{i\in [n]}\exp(\lambda X_i)\right]\cr&\le\sum_{i=1}^n\mathbb{E}[\exp(\lambda X_i)]\cr&\le n\exp(\lambda\mu+\lambda^2(\sigma^{\star})^2/2).
\end{align*}
 Set $\lambda=\frac{\sqrt{2\log( n)}}{\sigma^{\star}},$ then we get $$\mathbb{E}[X^\star]\le\mu+ \sigma^{\star} \sqrt{2\log(n)}.$$
\end{proof}
Denote by $Var(X)$ the variance of a random variable $X$ and $V^{\star}$ the maximum value among $Var(\overline{x}(\nu_f,\Sigma,x_{a,t})^\top\theta^\prime)$ for all $a\in\mathcal{A}$. Then using Lemma~\ref{lem:maxgaussian} and $\A_t \subseteq \mathcal{A}$, we have
\begin{align}
&\mathbb{E} \left[ \overline{x}(\nu_f,\Sigma,x_{a^\star_t,t})^\top 
  \theta^\prime - \overline{x}(\nu_f,\Sigma,x_{a_t,t})^\top
  \theta^\prime\right]\cr
  &\le \mathbb{E} \left[ \max_{a\in \mathcal{A}}\left( \overline{x}(\nu_f,\Sigma,x_{a,t})^\top\theta^\prime\right)+ \max_{a\in \mathcal{A}}\left(-
     \overline{x}(\nu_f,\Sigma,x_{a,t})^\top \theta^\prime\right)
     \right] \cr
     &\le \mathbb{E}[ 2\sqrt{2V^{\star}\log (K)}], 
    \label{eq:r_max}
\end{align}
where $V^{\star}$ is bounded as follows. Define $\mathcal{E}=\{1,...,d\}$. For any $a \in \mathcal{A}$ and any given $m_{a,t}$, we have
\begin{align*}
      Var(\overline{x}(\nu_f,\Sigma,x_{a,t})^\top\theta^\prime)
    &=Var((\nu_f+\Sigma_{\mathcal{ES}}\Sigma_{\mathcal{SS}}^{-1}(x_{a,t}-\nu_f)_\mathcal{S})^\top(\Sigma^{-1}\Sigma_f\theta^{\star}))\cr
    &=Var((\Sigma_{\mathcal{ES}}\Sigma_{\mathcal{SS}}^{-1}(x_{a,t})_\mathcal{S})^\top\Sigma^{-1}\Sigma_f\theta^{\star})\cr
    &=\mathbb{E}\left[((x_{a,t}-\nu_f)_\mathcal{S}^\top\Sigma_{\mathcal{SS}}^{-1}\Sigma_{\mathcal{ES}}^\top\Sigma^{-1}\Sigma_f\theta^{\star})^2\right]\cr
    &=O\left({\theta^{\star}}^\top\Sigma_f\Sigma^{-1}\Sigma_{\mathcal{ES}}\Sigma_{\mathcal{SS}}^{-1}\Sigma_{\mathcal{ES}}^\top\Sigma^{-1}\Sigma_f\theta^{\star}\right)\cr
    &=O\left(\|\theta^{\star}\|_2^2\|\Sigma_f\|_2^2\|\Sigma^{-1}\|_2^2\|\Sigma_{\mathcal{ES}}\|_2^2\|\Sigma_{\mathcal{SS}}^{-1}\|_2\right)\cr
    &=O\left( \|\theta^{\star}\|_2^2\|\Sigma\|_2^4\|\Sigma^{-1}\|_2^2\|\Sigma_{\mathcal{SS}}^{-1}\|_2\right)\cr
    &=O\left(1\right).
\end{align*}
\end{proof}

Therefore from Lemma~\ref{lem:max_r}, the regret over the first $\tau$ time steps in \eqref{eq:overRT} is replaced by
\begin{align}
\sum_{t=1}^{\tau} \mathbb{E} \left[\overline{x}(\nu_f,\Sigma,x_{a^\star_t,t})^\top \theta^\prime - \overline{x}(\nu_f,\Sigma,x_{a_t,t})^\top \theta^\prime\right] &=  O\left(\tau\sqrt{\log (K)} \right).\label{eq:exploration_regret} \end{align}

Now we provide a regret bound after $\tau$ time steps. Recall that instantaneous regret $r_t= \overline{z}_{a^\star_t,t}^\top\overline{\theta}- \overline{z}_{a_t,t}^\top\overline{\theta}$. In the algorithm, at each time step, the covariance matrix $\Sigma$ is estimated by the scaled empirical covariance matrix $\widehat{\Sigma}$ and the mean of feature vector $\nu_f$ is estimated by $\widehat{\nu}$. For ease of presentation, we write $\widehat{p}_t$, $\widehat{\nu}_t$, and $\widehat{\Sigma}_t$ for $\widehat{p}$, $\widehat{\nu}$, and $\widehat{\Sigma}$ at time step $t$, respectively. 
For analyzing the regret, we can decompose the regret according to the estimation errors with high probability as follows:
\begin{align*}
    r_t\le \|\overline{z}_{a^\star_t,t}-\widehat{z}_{a^\star_t,t}(t)\|_2\| \overline{\theta}\|_2+\|\tilde{\theta}_t - \overline{\theta}\|_{V_{t-1}}\|\widehat{z}_{a_t,t}(t)\|_{V_{t-1}^{-1}} +\|\widehat{z}_{a_t,t}(t) -   \overline{z}_{a_t,t}\|_2\| \overline{\theta}\|_2,
\end{align*}
which will be shown later.
The instantaneous regret bound consists of the estimation errors of $\widehat{z}_{a_t^\star,t}(t)$, $\widehat{z}_{a_t,t}(t)$, and $\tilde{\theta}_t$.
Therefore we first focus on providing bounds for $\| \overline{z}_{a,t}-\widehat{z}_{a,t}(t) \|_2$ and $\|\overline{\theta}-\tilde{\theta}_t\|_2.$
From matrix concentration inequalities, we have the following lemma. 

\begin{lemma}
  For all $t>\tau/2$, with probability at least $1-3/T$, we have
$$\left\|\nu_f - \widehat{\nu}_t \right\|_2= O\left(\frac{1}{p}\sqrt{\frac{d\log(T)}{tK}}\right), \left\|\Sigma - \widehat{\Sigma}_t \right\|_2= O\left(\frac{1}{p^2}\sqrt{\frac{d\log(T)}{tK}}\right),\mbox{ and } |\widehat{p} _t- p | = O\left( \sqrt{\frac{\log (T)}{dtK }} \right) $$
\label{lem:sampletheta}
\end{lemma}
\begin{proof}
Recall $x^\prime_{a,t}=z_{a,t}+\varepsilon_{a,t}$. We define $v_{a,t} = x^\prime_{a,t}\circ m_{a,t}-\nu_fp$ for all $a\in\mathcal{A}$. Let $z_j$ be the $j$-th entry in an arbitrary vector $z$. Then, for any $u\in\mathbb{R}^{d}$ such that $\|u\|_2=1$ and for any $\lambda \in \mathbb{R}$, we get
\begin{align*}
    &\mathbb{E}[\exp (\lambda v_{a,t}^\top u)]\cr &=\mathbb{E}\left[\mathbb{E}\left[\exp\left(\lambda ((x^\prime_{a,t}-\nu_f)\circ m_{a,t})^\top u\right)|m_{a,t}\right]\exp(\lambda(\nu_f\circ m_{a,t}-\nu_fp)^\top u)\right] \\
    &\le\exp(\|\Sigma\|_2\lambda^2/2)\mathbb{E}\left[\exp(\lambda(\nu_f\circ m_{a,t}-\nu_fp)^\top u)\right]\\
    &\le \exp(\|\Sigma\|_2\lambda^2/2)\prod_{j=1}^d\mathbb{E}\left[\exp(\lambda(\nu_f\circ m_{a,t}-\nu_fp)_j u_j)\right]\\
    &\le \exp(\|\Sigma\|_2\lambda^2/2)\exp(\sum\limits_{j=1}^d|{(\nu_f)}_j u_j|^2\lambda^2/2 )\\
    &\le \exp(\|\Sigma\|_2\lambda^2/2)\exp(\sum\limits_{j=1}^d|{(\nu_f)}_j u_j|^2\lambda^2/2 )\\
    &\le \exp((\|\Sigma\|_2+1)\lambda^2/2),
\end{align*}
where the first inequality is from the normal moment generating function and the third inequality is from Hoeffding's lemma. 
From the definition of sub-gaussian vector, we can find that $v_{a,t}$ for $a\in \mathcal{A}$ and $t>0$ are according to independent sub-gaussian with variance proxy $\|\Sigma\|_2+1$.
According to Theorem 2.1 of \cite{hsu2012tail}, for all $k>0$ we get 
\begin{align}
    \mathbb{P}\left(\left\|\sum_{s=1}^t\sum_{a\in\mathcal{A}} (x^\prime_{a,s}\circ m_{a,s}-\nu_fp)\right\|_2\ge \sqrt{tK(\|\Sigma\|_2+1)\left(d+2\sqrt{dk}+2k\right)}\right)\le \exp{(-k)}. 
    \label{eq:xsubGau}
\end{align}
Then we have with probability at least $1-1/T^2$ $$\left\|\frac{1}{t K}\sum_{s=1}^t\sum_{a\in\mathcal{A}}(x_{a,s}-p\nu_f)\right\|_2=O\left(\sqrt{\frac{d+\log(T)+\sqrt{d\log(T)}}{tK}}\right)$$
 By Hoeffding's inequality, we also have 
\begin{equation}|\widehat{p} _t- p | = O\left( \sqrt{\frac{\log (T)}{dtK }} \right) \quad \mbox{w.p. at least } 1-1/T^2.\label{eq:hoeff}\end{equation}
Recall that $\tau/2=\lceil d(\log(T))^2/(Kp^4)\rceil$. Then for $t>\tau/2$, we have $\widehat{p}_t=\Theta(p)$ at least probability $1-1/T^2$.
Using the above inequalities, we provide bounds for  $\|\nu_f-\widehat{\nu}_t\|_2$ and $\|\Sigma - \widehat{\Sigma}_t\|_2$. With probability at least $1-1/T^2$, for $t>\tau/2$  we have
\begin{align*}
    \|\widehat{\nu}_t-\nu_f\|_2&=\left\|\widehat{\nu}_t-\frac{1}{tKp}\sum_{s=1}^{t}\sum_{a\in\mathcal{A}}x_{a,s}+\frac{1}{tKp}\sum_{s=1}^{t}\sum_{a\in\mathcal{A}}x_{a,s}-\nu_f\right\|_2\\
    &\le \left|\frac{p-\widehat{p}_t}{p\widehat{p}_t}\right|\left(\left\|\frac{1}{tK}\sum_{s=1}^{t}\sum_{a\in\mathcal{A}}(x_{a,s}-p\nu_f)\right\|_2+\|p\nu_f\|_2\right)+\left\|\frac{1}{tKp}\sum_{s=1}^{t}\sum_{a\in\mathcal{A}}x_{a,s}-\nu_f\right\|_2\\
    &=O\left(\frac{1}{p}\sqrt{\frac{d\log(T)}{tK}}\right).
    \end{align*}
    
    Now we provide a bound for $\|\Sigma-\widehat{\Sigma}_t\|_2$.
 We define $w=
\nu_fp$. By using $x_{a,t}=v_{a,t}+w$, we get
\begin{align}
    \left\|\sum_{s=1}^t\sum_{a\in\mathcal{A}} x_{a,s}x_{a,s}^\top  -\mathbb{E} \left[ \sum_{s=1}^t\sum_{a\in\mathcal{A}} x_{a,s}x_{a,s}^\top \right] \right\|_2
    &\nonumber\le
    \left\|\sum_{s=1}^t\sum_{a\in\mathcal{A}} v_{a,s}v_{a,s}^\top - \mathbb{E}\left[\sum_{s=1}^t\sum_{a\in\mathcal{A}} v_{a,s}v_{a,s}^\top\right]\right\|_2\\
    &\quad+2\left\|\sum_{s=1}^t\sum_{a\in\mathcal{A}} v_{a,s}w^\top - \mathbb{E}\left[\sum_{s=1}^t\sum_{a\in\mathcal{A}} v_{a,s}w^\top\right]\right\|_2 \label{eq:vvT}.    
\end{align}
For the first term in the RHS of \eqref{eq:vvT}, from Proposition 2.1 of \cite{vershynin2012close} with sub-Gaussian $v_a$, with probability at least $1-1/T^2$ we get
\begin{align}
\left\| \sum_{s=1}^t\sum_{a\in\mathcal{A}} v_{a,s}v_{a,s}^\top  -\mathbb{E} \left[ \sum_{s=1}^t\sum_{a\in\mathcal{A}} v_{a,s}v_{a,s}^\top \right] \right\|_2 
 = O\left(\sqrt{t K d \log(T)}\right). \label{eq:v1v1T}
\end{align}

For the second term in the RHS of \eqref{eq:vvT}, 

\begin{align}
\left\| \sum_{s=1}^t\sum_{a\in\mathcal{A}} v_{a,s}w^\top  -\mathbb{E} \left[ \sum_{s=1}^t\sum_{a\in\mathcal{A}} v_{a,s}w^\top \right] \right\|_2 
\nonumber&=\left\| \sum_{s=1}^t\sum_{a\in\mathcal{A}} v_{a,s}w^\top \right\|_2\\
&\le\left\| \sum_{s=1}^t\sum_{a\in\mathcal{A}}\left(  x^\prime_{a,s}\circ m_{a,s}-\nu_fp\right) \right\|_2\|\nu_fp\|_2
.\label{eq:v1v2Tbound}
\end{align}

From the results of \eqref{eq:v1v2Tbound} and \eqref{eq:xsubGau}, with probability at least $1-1/T^2$ we get
\begin{align}
    \left\| \sum_{s=1}^t\sum_{a\in\mathcal{A}} v_{a,s}w^\top  -\mathbb{E} \left[ \sum_{s=1}^t\sum_{a\in\mathcal{A}} v_{a,s}w^\top \right] \right\|_2 &= O\left(\left\|\sum_{s=1}^t\sum_{a\in\mathcal{A}}(x^\prime_{a,s}\circ m_{a,s}-\nu_fp)\right\|_2p\right)\cr
    &=O\left(p\sqrt{tK d\ln(T)}\right).\label{eq:v1v2TboundO}
\end{align}

By putting the results of \eqref{eq:vvT}, \eqref{eq:v1v1T}, and \eqref{eq:v1v2TboundO}, with probability at least $1-1/T^2$ we get 
\begin{align}
\left\| \sum_{s=1}^t\sum_{a\in\mathcal{A}} x_{a,s}x_{a,s}^\top  -\mathbb{E} \left[ \sum_{s=1}^t\sum_{a\in\mathcal{A}} x_{a,s}x_{a,s}^\top \right] \right\|_2 
&=O\left(\sqrt{tK d \log(T)}\right).
\label{eq:vvTO}
\end{align}
We write $Z_t$ for $Z$ at time step $t$ in the algorithm. Then, \eqref{eq:vvTO} directly implies that
\begin{align}
\left\|Z_t - \mathbb{E}[Z_t]  \right\|_2=O\left( \sqrt{tK d \log(T)}\right)\label{eq:XEX2} .\end{align}
Let $\textbf{1}_{d\times d}$ be a matrix $\in\mathbb{R}^{d\times d}$ with all $1$ entries and $P=(p-p^2)I_d+p^2{\bm 1_{d\times d}}$. Then from Proposition 1 in \cite{pavez2020covariance}, we can show that 
\begin{align}
    \|\mathbb{E}[Z_t]\|_2=tK\|\Sigma\circ P+(\nu\nu^\top) \circ P\|_2=O(tKp) \label{eq:EX2}
\end{align}
 Lastly, with probability at least $1-1/T^2$, we get
\begin{align*}
    \|\widehat{\Sigma}_t-\Sigma\|_2 &\le \left\|\widehat{\Sigma}_t-\left(\frac{1}{tK}Z_t\circ\left(\frac{p-1}{p^2}I_{d}+\frac{1}{p^2}\textbf{1}_{d\times d}\right)-\nu_f\nu_f^\top\right)\right.\cr&\left.\qquad+\left(\frac{1}{tK}Z_t\circ\left(\frac{p-1}{p^2}I_{d}+\frac{1}{p^2}\textbf{1}_{d\times d}\right)-\nu_f\nu_f^\top\right)-\Sigma\right\|_2\cr
    &\le \left\|\widehat{\Sigma}_t-\left(\frac{1}{tK}Z_t\circ\left(\frac{p-1}{p^2}I_{d}+\frac{1}{p^2}\textbf{1}_{d\times d}\right)-\nu_f\nu_f^\top\right)\right\|_2\cr& \qquad+\left\|\left(\frac{1}{tK}Z_t\circ\left(\frac{p-1}{p^2}I_{d}+\frac{1}{p^2}\textbf{1}_{d\times d}\right)-\nu_f\nu_f^\top\right)-\Sigma\right\|_2\cr
    &= O\left(\frac{1}{tK}\frac{|p-\widehat{p}_t|}{p^3}(\|Z_t-\mathbb{E}[Z_t]\|_2+\|\mathbb{E}[Z_t]\|_2)+\|\widehat{\nu}_t-\nu_f\|_2\right.\cr &\left.\qquad+\frac{1}{t K}\left\|(Z_t-\mathbb{E}[Z_t])\circ\left(\frac{p-1}{p^2}I_{d}+\frac{1}{p^2}\textbf{1}_{d\times d}\right)\right\|_2\right)\cr
    &= O\left(\frac{1}{p^2}\sqrt{\frac{d\log(T)}{tK}}\right).
\end{align*}
Therefore, using the union bound for all time $t>\tau/2$, we can conclude the proof.
\end{proof}
  From Lemma~\ref{lem:sampletheta}, we define an event \begin{align}E_1 &=\left\{\left\|\nu_f - \widehat{\nu}_t \right\|_2= O\left(\frac{1}{p}\sqrt{\frac{d\log(T)}{tK}}\right),  \left\|\Sigma - \widehat{\Sigma}_t \right\|_2= O\left(\frac{1}{p^2}\sqrt{\frac{d\log(T)}{tK}}\right),\right. \cr & \left.\qquad\qquad \mbox{ and } |\widehat{p} _t- p | = O\left( \sqrt{\frac{\log (T)}{dtK }}\right)  \forall t\in[\tau/2+1,T] \right\},\end{align} which holds true with at least probability $1-3/T$. 
  Using Weyl's inequality, we can show that
$\sigma_d(\widehat{\Sigma}_t)\ge \sigma_d(\Sigma)-\|\widehat{\Sigma}_t-\Sigma\|_2$.
Under $E_1$, for all $t>\tau/2$, we have $\|\Sigma-\widehat{\Sigma}_t\|_2=o(1).$ Then, we have
\begin{gather}
    \|\widehat{\Sigma}_t\|_2\le \|\widehat{\Sigma}_t-\Sigma\|_2+\|\Sigma\|_2=O\left(\|\Sigma\|_2\right) \quad\text{ and}
    \cr
    \|\widehat{\Sigma}_t^{-1}\|_2=\frac{1}{\sigma_d(\widehat{\Sigma}_t)}\le\frac{1}{\sigma_d(\Sigma)-\|\widehat{\Sigma}_t-\Sigma\|_2}=O\left(\frac{1}{\sigma_d(\Sigma)}\right)=O\left(\|\Sigma^{-1}\|_2\right).\label{eq:SigBig}
\end{gather} 
By following the same steps, for any non empty set $\mathcal{S}$, we can also show that 
\begin{align}
    \|(\widehat{\Sigma}_t)^{-1}_{\mathcal{S}\mathcal{S}}\|_2=O(\|\Sigma_{\mathcal{S}\mathcal{S}}^{-1}\|_2).\label{eq:sub_SigBig}
\end{align}

Then, under $E_1$, from the definition of $\widehat{z}_{a,t}(t)$ and \eqref{eq:SigBig}, for any $t> \tau$ and $0<s\le t$ we have
\begin{align*}
    \|\widehat{z}_{a_s,s}(t)\|_2=O(\sqrt{d\log(KT)}).
\end{align*}
This is because the algorithm updates the estimated features of previously chosen arms once within $\tau/2<t\le\tau$.
  In what follows, we provide a bound for feature estimators using \eqref{eq:SigBig}. For simplicity, we define $\overline{z}_{a,t}\coloneqq[1;\overline{x}(\nu,\Sigma,x_{a,t})]$. In the algorithm, recall that for $s\le t$, $\widehat{z}_{a,s}(t)=[1;\overline{x}(\widehat{\nu},\widehat{\Sigma},x_{a,s})]$ at time $t$. 
\begin{lemma}
Under $E_1$, for all $t>\tau/2$ and $a\in\A_t$, we have
\begin{align*}
 & \|\overline{z}_{a,t} - \widehat{z}_{a,t}(t)\|_2  =O\left(d\sqrt{\frac{1}{p^3}\frac{\log(KT)\log(T)}{tK}}\right). 
\end{align*}\label{lem:esterror2}
\end{lemma}
\begin{proof}

First, at time step $t$, we have
\begin{align}
&\Sigma_{\mathcal{U}\mathcal{S}}\Sigma_{\mathcal{S}\mathcal{S}}^{-1}\cr
 = \,& \Sigma_{\mathcal{U}\mathcal{S}}{(\widehat{\Sigma}_t)}_{\mathcal{S}\mathcal{S}}^{-1}{(\widehat{\Sigma}_t)}_{\mathcal{S}\mathcal{S}}\Sigma_{\mathcal{S}\mathcal{S}}^{-1}\cr
 =\, &\Sigma_{\mathcal{U}\mathcal{S}}{(\widehat{\Sigma}_t)}_{\mathcal{S}\mathcal{S}}^{-1}+ \Sigma_{\mathcal{U}\mathcal{S}}{(\widehat{\Sigma}_t)}_{\mathcal{S}\mathcal{S}}^{-1}({(\widehat{\Sigma}_t)}_{\mathcal{S}\mathcal{S}} - \Sigma_{\mathcal{S}\mathcal{S}})\Sigma_{\mathcal{S}\mathcal{S}}^{-1} \cr
 = \,&{(\widehat{\Sigma}_t)}_{\mathcal{U}\mathcal{S}}{(\widehat{\Sigma}_t)}_{\mathcal{S}\mathcal{S}}^{-1} + (\Sigma_{\mathcal{U}\mathcal{S}}-{(\widehat{\Sigma}_t)}_{\mathcal{U}\mathcal{S}}){(\widehat{\Sigma}_t)}_{\mathcal{S}\mathcal{S}}^{-1}+ \Sigma_{\mathcal{U}\mathcal{S}}{(\widehat{\Sigma}_t)}_{\mathcal{S}\mathcal{S}}^{-1}({(\widehat{\Sigma}_t)}_{\mathcal{S}\mathcal{S}} - \Sigma_{\mathcal{S}\mathcal{S}})\Sigma_{\mathcal{S}\mathcal{S}}^{-1}.\label{eq:SigmaSigmainv}
\end{align}

Thus with \eqref{eq:SigmaSigmainv}, \eqref{eq:SigBig}, and \eqref{eq:sub_SigBig}, under $E_1$ for $t>\tau/2$ 
\begin{align*}
&\left\| \Sigma_{\mathcal{U}\mathcal{S}}\Sigma_{\mathcal{S}\mathcal{S}}^{-1} - {(\widehat{\Sigma}_t)}_{\mathcal{U}\mathcal{S}}{(\widehat{\Sigma}_t)}_{\mathcal{S}\mathcal{S}}^{-1} \right\|_2\cr&
\le \,  \left\| (\Sigma_{\mathcal{U}\mathcal{S}}-{(\widehat{\Sigma}_t)}_{\mathcal{U}\mathcal{S}}){(\widehat{\Sigma}_t)}_{\mathcal{S}\mathcal{S}}^{-1} \right\|_2 + \left\| \Sigma_{\mathcal{U}\mathcal{S}}{(\widehat{\Sigma}_t)}_{\mathcal{S}\mathcal{S}}^{-1}({(\widehat{\Sigma}_t)}_{\mathcal{S}\mathcal{S}} - \Sigma_{\mathcal{S}\mathcal{S}})\Sigma_{\mathcal{S}\mathcal{S}}^{-1} \right\|_2 \cr&
= \, O\left( \|\Sigma - \widehat{\Sigma}_t \|_2\|\Sigma^{-1}_{\mathcal{S}\mathcal{S}}\|_2+\|\Sigma\|_2\|\Sigma_{\mathcal{S}\mathcal{S}}^{-1}\|_2\|\widehat{\Sigma}_t-\Sigma\|_2\|\Sigma^{-1}_{\mathcal{S}\mathcal{S}}\|_2\right) \cr&
= \, O\left(\|\widehat{\Sigma}_t-\Sigma\|_2\right).
\end{align*}

Therefore, for any $t>\tau/2$ and $a\in\mathcal{A}_t,$ it follows that under $E_1$,
\begin{align*}
\left\| \widehat{z}_{a,t}(t) - \overline{z}_{a,t}\right\|_2^2 &
\le 2\left\| \Sigma_{\mathcal{U}\mathcal{S}}\Sigma_{\mathcal{S}\mathcal{S}}^{-1} - {(\widehat{\Sigma}_t)}_{\mathcal{U}\mathcal{S}}{(\widehat{\Sigma}_t)}_{\mathcal{S}\mathcal{S}}^{-1} \right\|_2^2 \left\|(x_{a,t}-\nu_f)_\mathcal{S}\right\|_2^2 \cr & \qquad + 2\left(\|\nu_f-\widehat{\nu}_t \|_2 + \|{(\widehat{\Sigma}_t)}_{\mathcal{U}\mathcal{S}}{(\widehat{\Sigma}_t)}_{\mathcal{S}\mathcal{S}}^{-1}(\nu_f-\widehat{\nu}_t)_{\mathcal{S}}\|_2  \right)^2 \cr
 &= O\left( \|\Sigma - \widehat{\Sigma}_t \|_2^2\mathbb{E}[\|m_{a,t}\|_2^2]\log(KT) + \|\nu_f - \widehat{\nu}_t \|_2^2\right) \cr
 &=O\left(pd\|\Sigma - \widehat{\Sigma}_t \|_2^2\log(KT) + \|\nu_f - \widehat{\nu}_t \|_2^2\right)\cr
 &=O\left(d^2(1/p^3)\log(T)\log(KT)/tK\right).
\end{align*}
\end{proof}

Now, we provide a lemma for showing a good property of the confidence set $\mathcal{C}_t$. 
\begin{lemma}
Under $E_1$, for all $t > \tau$, with probability at least $1-1/T$, we have $\overline{\theta}\in \mathcal{C}_{t-1}.$\label{lem:confi}
\end{lemma}
\begin{proof} Since, under $E_1$,  $\|\widehat{\Sigma}\|_2=O(\|\Sigma\|_2)$ and $\|\widehat{\Sigma}^{-1}\|_2=O(\|\Sigma^{-1}\|_2)$, 
we have, for all $t>\tau$ and $1\le s\le t$ considering rarely updating estimators in the algorithm, $$\|\widehat{z}_{a_s,s}(t)\|_2=O(\sqrt{d\log(KT)}).$$
This is because the algorithm updates the estimated features of previously chosen arms once within $\tau/2<t\le\tau$. Then for $t>\tau$ we can observe that  $\|\widehat{z}_{a_s,s}(t)\|_2^2\le C_1d\log(KT)$ and $\|\overline{\theta}\|_2\le C_2$ for some constants $C_1$ and $C_2>0$. We also observe that noise of reward $\eta_t$ is independent to $E_1$. Using the facts, Lemma~\ref{lem:esterror2}, and Theorem 2 in \cite{abbasi2011improved}, we can easily prove that for all $t>\tau$, with probability at least $1-1/T$, $\overline{\theta}$ lies in $\mathcal{C}_{t-1}$. In what follows, we provide details. From Section 3.1 in \citet{smith2006optimizer}, Proposition 3.13 in \citet{eaton1983multivariate}, and $1$-subGaussian $\eta_t$, we observe that $y_{a_t}$ given $x_{a,t}$ follows a $C_3$-subGaussian distribution with mean $\overline{x}^\top \overline{\theta}$ for some constant $C_3>0$. We denote by $\tilde{\eta}_t$ the noise term for $y_{a_t}$ given $x_{a,t}$ at time $t$. Then, we have

\begin{align*}
    \widehat{\theta}_t&=V_t^{-1}\sum_{s=1}^t\widehat{z}_{a_s,s}(t+1)\left(\overline{z}_{a_s,s}^\top\overline{\theta}+\tilde{\eta}_s\right)\cr 
    &=V_t^{-1}\sum_{s=1}^t\widehat{z}_{a_s,s}(t+1)\left(\widehat{z}_{a_s,s}(t+1)^\top\overline{\theta}+\tilde{\eta}_s\right)+V_t^{-1}\sum_{s=1}^t\widehat{z}_{a_s,s}(t+1)(\overline{z}_{a_s,s}-\widehat{z}_{a_s,s}(t+1))^\top\overline{\theta}\cr
    &=V_t^{-1}\sum_{s=1}^t \widehat{z}_{a_s,s}(t+1)\tilde{\eta}_s+\overline{\theta}-\lambda  V_t^{-1}\overline{\theta}+V_t^{-1}\sum_{s=1}^t\widehat{z}_{a_s,s}(t+1)(\overline{z}_{a_s,s}-\widehat{z}_{a_s,s}(t+1))^\top\overline{\theta}.
    \end{align*}
Then $\forall x\in\mathbb{R}^{d+1}$, we have
\begin{align*}
|x^\top \widehat{\theta}_t-x^\top\overline{\theta}|&\le \|x\|_{V_t^{-1}}\left(\left\|\sum_{s=1}^t\widehat{z}_{a_s,s}(t+1)\tilde{\eta}_s\right\|_{V_t^{-1}}+\lambda^{1/2}\|\overline{\theta}\|_2+\left\|\sum_{s=1}^t\widehat{z}_{a_s,s}(t+1)(\overline{z}_{a_s,s}-\widehat{z}_{a_s,s}(t+1))^\top \overline{\theta}\right\|_{V_t^{-1}}\right)\cr 
&\le \|x\|_{V_t^{-1}}\left(\left\|\sum_{s=1}^t\widehat{z}_{a_s,s}(t+1)\tilde{\eta}_s\right\|_{V_t^{-1}}+\lambda^{1/2}\|\overline{\theta}\|_2+\sum_{s=1}^t\left\|\widehat{z}_{a_s,s}(t+1)\right\|_{V_t^{-1}}\|\overline{z}_{a_s,s}-\widehat{z}_{a_s,s}(t+1)\|_2 \|\overline{\theta}\|_2\right).
\end{align*}
By setting $x=V_t(\widehat{\theta}_t-\overline{\theta})$, we have
\begin{align*}
    \|\widehat{\theta}_t-\overline{\theta}\|_{V_t}\le \left\|\sum_{s=1}^t\widehat{z}_{a_s,s}(t+1)\tilde{\eta}_s\right\|_{V_t^{-1}}+\lambda^{1/2}\|\overline{\theta}\|_2+\sum_{s=1}^t\left\|\widehat{z}_{a_s,s}(t+1)\right\|_{V_t^{-1}}\|\overline{z}_{a_s,s}-\widehat{z}_{a_s,s}(t+1)\|_2 \|\overline{\theta}\|_2.
\end{align*}
From Lemma 9 in \citet{abbasi2011improved},  with probability at least $1-1/T$ we have
\begin{align*}
    \left\|\sum_{s=1}^t\widehat{z}_{a_s,s}(t+1)\tilde{\eta}_s\right\|_{V_t^{-1}}\le C_3\sqrt{2(d+1)\log(T(t+1))}.
\end{align*}
We can observe that $\widehat{p}_t=\Theta(p)$ when $t>\tau$ under $E_1$.
Putting them together, for some constant $C_5,C_6>0$, with probability at least $1-1/T$, we have
\begin{align*}
\|\widehat{\theta}_t-\overline{\theta}\|_{V_t}&\le C_5\sqrt{(d+1)\log(T(t+1))}+C_2\lambda^{1/2}+\sum_{s=1}^t\left\|\widehat{z}_{a_s,s}(t+1)\right\|_{V_t^{-1}}\frac{d}{p^{3/2}}\sqrt{\frac{\log(T)\log(KT)}{(t+1)K}}\cr 
&\le C_5\sqrt{(d+1)\log(T(t+1))}+C_2\lambda^{1/2}+C_6\sum_{s=1}^t\left\|\widehat{z}_{a_s,s}(t+1)\right\|_{V_t^{-1}}\frac{d}{\widehat{p}^{3/2}}\sqrt{\frac{\log(T)\log(KT)}{(t+1)K}}.
\end{align*}

\end{proof}
 From the above lemma, we define an event $$E_2=\left\{\overline{\theta}\in \mathcal{C}_{t-1},  \forall t\in[\tau+1,T]\right\}.$$
 Then we provide a lemma for decomposing the instantaneous regret according to the estimation errors of $\widehat{z}_{a_t^\star,t}(t)$, $\widehat{z}_{a_t,t}(t)$, and $\tilde{\theta}_t$.
 \begin{lemma} Under $E_2$,
we have 
\begin{align*}
    r_t\le \|\overline{z}_{a^\star_t,t}-\widehat{z}_{a^\star_t,t}(t)\|_2\| \overline{\theta}\|_2+\|\tilde{\theta}_t - \overline{\theta}\|_{V_{t-1}}\|\widehat{z}_{a_t,t}(t)\|_{V_{t-1}^{-1}} +\|\widehat{z}_{a_t,t}(t) -   \overline{z}_{a_t,t}\|_2\| \overline{\theta}\|_2
\end{align*}\label{lem:regret_errors}
\end{lemma}
\begin{proof} Under $E_2$, from the fact that $\overline{\theta}\in \mathcal{C}_{t-1}$ and considering that $a_t$ is the chosen arm at time $t$, we have $ \widehat{z}_{a_t^\star,t}(t)^\top\overline{\theta}\le\widehat{z}_{a_t,t}(t)^\top\tilde{\theta}_t$. Then we have
\begin{align*}
    r_t&= \overline{z}_{a^\star_t,t}^\top\overline{\theta}- \overline{z}_{a_t,t}^\top\overline{\theta}\cr
    &=\overline{z}_{a^\star_t,t}^\top\overline{\theta}-\widehat{z}_{a^\star_t,t}(t)^\top \overline{\theta}+\widehat{z}_{a^\star_t,t}(t)^\top \overline{\theta}- \overline{z}_{a_t,t}^\top\overline{\theta}\cr
    &\le\|\overline{z}_{a^\star_t,t}-\widehat{z}_{a^\star_t,t}(t)\|_2\| \overline{\theta}\|_2+\widehat{z}_{a^\star_t,t}(t)^\top \overline{\theta}- \overline{z}_{a_t,t}^\top\overline{\theta}\cr
    &\le \|\overline{z}_{a^\star_t,t}-\widehat{z}_{a^\star_t,t}(t)\|_2\| \overline{\theta}\|_2+\widehat{z}_{a_t,t}(t)^\top\tilde{\theta}_t -  \overline{z}_{a_t,t}^\top \overline{\theta} \cr
    &\le \|\overline{z}_{a^\star_t,t}-\widehat{z}_{a^\star_t,t}(t)\|_2\| \overline{\theta}\|_2+\widehat{z}_{a_t,t}(t)^\top\tilde{\theta}_t - \widehat{z}_{a_t,t}(t)^\top\overline{\theta} +\widehat{z}_{a_t,t}(t)\overline{\theta} -   \overline{z}_{a_t,t} \overline{\theta} \cr
        &\le \|\overline{z}_{a^\star_t,t}-\widehat{z}_{a^\star_t,t}(t)\|_2\| \overline{\theta}\|_2+\|\tilde{\theta}_t - \overline{\theta}\|_{V_{t-1}}\|\widehat{z}_{a_t,t}(t)\|_{V_{t-1}^{-1}} +\|\widehat{z}_{a_t,t}(t) -   \overline{z}_{a_t,t}\|_2\| \overline{\theta}\|_2.
\end{align*}
\end{proof}
Then, we provide a bound for $\sum_{t=\tau+1}^T\mathbb{E}[r_t]$ in the following. 
\begin{lemma} We have
\begin{align}
     \sum_{t=\tau+1}^T\mathbb{E}[r_t] &=\tilde{O}\left(\sqrt{d}\mathbb{E}\left[\sum_{t=\tau+1}^T\|\widehat{z}_{a_t,t}(t)\|_{V_{t-1}^{-1}}\mid E_1\cap E_2\right]+\frac{d}{p^{3/2}}\sqrt{\frac{T}{K}}+\sqrt{d}\right).\label{eq:r_bd_2}\cr
\end{align}\label{lem:r_sum}\end{lemma}
\begin{proof}
Under $E_1$ and $E_2$, we have $\tilde{\theta}_t\in\mathcal{C}_{t-1}$. Then we get
\begin{align*}
\|\tilde{\theta}_t-\overline{\theta}\|_{V_{t-1}} 
&\le C_5\sqrt{(d+1)\log(Tt))}+C_2\lambda^{1/2}+C_6\sum_{s=1}^{t-1}\left\|\widehat{z}_{a_s,s}(t)\right\|_{V_{t-1}^{-1}}\frac{d}{\widehat{p}^{3/2}}\sqrt{\frac{\log(T)\log(KT)}{tK}}
\cr
&\le C_5\sqrt{2(d+1)\log(Tt)}+C_2\lambda^{1/2}+C_6\sqrt{t\sum_{s=1}^{t-1}\left\|\widehat{z}_{a_s,s}(t)\right\|_{V_{s-1}^{-1}}^2}\frac{d}{\hat{p}^{3/2}}\sqrt{\frac{\log(T)\log(KT)}{tK}}\cr
&=O\left(\sqrt{d\log(Tt)}+\lambda^{1/2}+\frac{d^{3/2}}{p^{3/2}}\log(T)\sqrt{\frac{\log(KT)}{K}}\right),
\end{align*}
where the last equality comes from Lemma 11 in \citet{abbasi2011improved}.

Then with Lemma~\ref{lem:regret_errors}, we have
\begin{align}
    r_t&\le \|\overline{z}_{a^\star_t,t}-\widehat{z}_{a^\star_t,t}(t)\|_2\| \overline{\theta}\|_2+\|\tilde{\theta}_t - \overline{\theta}\|_{V_{t-1}}\|\widehat{z}_{a_t,t}(t)\|_{V_{t-1}^{-1}} +\|\widehat{z}_{a_t,t}(t) -   \overline{z}_{a_t,t}\|_2\| \overline{\theta}\|_2\cr
    &=\tilde{O}\left(\left(\sqrt{d}+(d/p)^{3/2}\sqrt{1/K}\right)\|\widehat{z}_{a_t,t}(t)\|_{V_{t-1}^{-1}}+\|\overline{z}_{a^\star_t,t}-\widehat{z}_{a^\star_t,t}(t)\|_2\| \overline{\theta}\|_2+\|\widehat{z}_{a_t,t}(t) -   \overline{z}_{a_t,t}\|_2\| \overline{\theta}\|_2\right),\cr\label{eq:r_bd}
\end{align}
where the equality comes from $\tilde{\theta}_t\in \mathcal{C}_{t-1}$.
Then from Lemma~\ref{lem:esterror2}, we have
\begin{align}
    r_t=\tilde{O}\left(\left(\sqrt{d}+(d/p)^{3/2}\sqrt{1/K}\right)\|\widehat{z}_{a_t,t}(t)\|_{V_{t-1}^{-1}}+(d/p^{3/2})\sqrt{1/tK}\right).
\end{align}
We note that under the complement event $E_1^c\cup E_2^c$ which holds at most probability $2/T$, for all $t>\tau$, we have  $r_t\le |\overline{z}_{a^\star_t,t}^\top\theta^\prime|+ |\overline{z}_{a_t,t}^\top\theta^\prime| =O(\sqrt{d\log(KT)}).$ Therefore with \eqref{eq:r_bd}, we have
\begin{align*}
    \sum_{t=\tau+1}^T\mathbb{E}[r_t]&=\tilde{O}\left(\left(\sqrt{d}+(d/p)^{3/2}\sqrt{1/K}\right)\mathbb{E}\left[\sum_{t=\tau +1}^T\|\widehat{z}_{a_t,t}(t)\|_{V_{t-1}^{-1}}\mid E_1\cap E_2\right]+(d/p^{3/2})\sqrt{T/K}+\sqrt{d}\right)
\end{align*}

\end{proof}
Now we provide a lemma to bound the first term in \eqref{eq:r_bd_2}.
\begin{lemma} Under $E_1$, we have
$$\sum_{t=\tau+1}^T\|\widehat{z}_{a_t,t}(t)\|_{V_{t-1}^{-1}}^2=O\left( d\log(T\log(KT))\log(T)\right)$$\label{lem:x_sum}
\end{lemma}
\begin{proof}

Under $E_1$, for some sufficiently large constant $C>0$ with $\lambda=Cd\log(KT)$, for any $t\ge\tau+1$ and $s\le t$ we have
$$\|\widehat{z}_{a_s,s}(t)\|_2^2\le Cd\log(KT) \text{ and }$$
$$\|\widehat{z}_{a_s,s}(t)\|_{V_{t-1}^{-1}}^2\le \|\widehat{z}_{a_s,s}(t)\|_2^2\|V_{t-1}^{-1}\|_2\le\|\widehat{z}_{a_s,s}(t)\|_2^2/Cd\log(KT)\le 1.$$
Let $2^l$ be the smallest time step after $\tau+1$ for an integer $l$ and $2^n$ be the largest time step before $T$ for an integer $n$. Then we have $n=O(\log T)$. We note that estimated features of previously selected arms in $V_{t-1}$ are not updated during the time steps between $t=2^i+1$ and $2^{i+1}-1$ for $i\in[n]$.
Then by following the proof steps in Lemma 11 in \citet{abbasi2011improved} and considering rarely updating procedure in the algorithm, we can obtain
\begin{align}
\sum_{t=\tau+1}^T\|\widehat{z}_{a_t,t}(t)\|_{V_{t-1}^{-1}}^2&=\sum_{t=\tau+1}^{2^l-1}\|\widehat{z}_{a_t,t}(t)\|_{V_{t-1}^{-1}}^2+ \sum_{i=l}^{n-1}\sum_{t=2^i}^{2^{i+1}-1}\|\widehat{z}_{a_t,t}(t)\|_{V_{t-1}^{-1}}^2+ \sum_{t=2^n}^{T}\|\widehat{z}_{a_t,t}(t)\|_{V_{t-1}^{-1}}^2\cr 
&\le 2\sum_{t=\tau+1}^{2^l-1}\log\left(1+\|\widehat{z}_{a_t,t}(t)\|_{V_{t-1}^{-1}}^2\right)\cr &\qquad+ 2\sum_{i=l}^{n-1}\sum_{t=2^i}^{2^{i+1}-1}\log\left(1+\|\widehat{z}_{a_t,t}(t)\|_{V_{t-1}^{-1}}^2\right)+ 2\sum_{t=2^n}^{T}\log\left(1+\|\widehat{z}_{a_t,t}(t)\|_{V_{t-1}^{-1}}^2\right)
\cr &\le 2(\log(\det(V_{2^l-1}))+\sum_{i=l}^{n-1}2(\log(\det(V_{2^{i+1}-1}))+2(\log(\det(V_{T}))\cr &
\le 2(n+2)(d+1)\log(C(T+1)\log(KT))\cr &=O\left(d\log(T\log(KT))\log(T)\right),
\end{align}
where the first inequality is obtained from $x\le 2\log(1+x)$ when $x\in[0,1]$, the second and last inequality is obtained from Lemma 11 in \citet{abbasi2011improved}, and the last equality is obtained from $n=O(\log T)$. 
\end{proof}

Finally  using \eqref{eq:overRT}, \eqref{eq:exploration_regret}, Lemmas~\ref{lem:r_sum} and \ref{lem:x_sum}, and Cauchy-Schwarz inequality, we get
\begin{align*}
    R(T)&=\mathbb{E}\left[\sum_{t=1}^T r_t\right]\cr
    &=\tilde{O}\left(\tau+\mathbb{E}\left[\left(\sqrt{d}+(d/p)^{3/2}\sqrt{1/K}\right)\sum_{t=\tau+1}^T\|\widehat{z}_{a_t,t}(t)\|_{V_{t-1}^{-1}}\mid E_1\cap E_2 \right]+\frac{d}{p^{3/2}}\sqrt{T/K}+\sqrt{d}\right)
    \cr
    &=\tilde{O}\left(d/Kp^4+\mathbb{E}\left[\left(\sqrt{d}+(d/p)^{3/2}\sqrt{1/K}\right)\sqrt{T\sum_{t=\tau+1}^T\|\widehat{z}_{a_t,t}(t)\|^2_{V_{t-1}^{-1}}}\mid E_1\cap E_2 \right]+\frac{d}{p^{3/2}}\sqrt{T/K}+\sqrt{d}\right)\cr &=\tilde{O}\left(\frac{d}{Kp^4}+d\sqrt{T}+\frac{d^2}{p^{3/2}}\sqrt{\frac{T}{K}}\right).
\end{align*}
\vspace{1mm}
\subsection{Tightness of the regret bound of Theorem~\ref{thm:regret} with respect to $p$}\label{app:p_tight}

 Dependency on $p$ in regret analysis comes from variance of estimator $\widehat{\Sigma}_t$ which includes $1/\widehat{p}_t^2$ term for reducing bias. In more detail,  from \eqref{eq:ins_reg} we can observe that regret depends on the estimation error for Bayesian feature estimator $\widehat{z}_a$, and this error depends on the variance of $\widehat{\Sigma}_t$ and $\|x_{a,t}\|_2$ (See proof of Lemma~\ref{lem:esterror2}). Then, from  Lemma~\ref{lem:sampletheta}, we get $\|\widehat{\Sigma}_t-\Sigma\|_2=O((1/p^2)\sqrt{d\log(T)/tK})$, and from Gaussian distribution we get $\|x_{a,t}\|_2=\tilde{O}(\sqrt{pd})$. Therefore, $1/p^{3/2}$ in the regret bound seems to be inevitable  considering $1/p^2$ in the variance of $\widehat{\Sigma}_t$ and $\sqrt{p}$ from $\|x_{a,t}\|_2$. 
 
 Next, we explain why $1/p^4$ appears in the regret bound. We can observe that the instantaneous regret depends on $\|\widehat{z}_{a_t}\|_2$ from \eqref{eq:r_t_bd} and Lemma~\ref{lem:x_sum}. For bounding the variance $\|\widehat{z}_{a_t}\|_2$, we need to bound $\|\widehat{\Sigma}^{-1}_t\|_2$ from the Bayesian feature definition.
 Using Weyl's inequality, we have $\|\widehat{\Sigma}^{-1}_t\|_2\le 1/(\sigma_d(\Sigma)-\|\widehat{\Sigma}_t-\Sigma\|_2)$ in \eqref{eq:SigBig}. Therefore, for getting a tight constant bound for $\|\widehat{\Sigma}^{-1}_t\|_2$, we need at least $t=\Theta(1/p^4)$ time steps to collect enough information for reducing $\|\widehat{\Sigma}_t-\Sigma_t\|_2$ from Lemma~\ref{lem:sampletheta}. This is why we have $1/p^4$ in the regret bound.
We believe that there may be some room to improve the dependency of $1/p^4$ by controlling $\|\widehat{\Sigma}_t^{-1}\|_2$, which can be a future work.

\subsection{Algorithm~\ref{alg:alg2} and its regret bound\label{app:eff_alg}}

\begin{algorithm}[t]
   \caption{Efficient Contextual Linear Bandits on Bayesian Features ({\tt E-CLBBF})}
    \label{alg:alg2} 
\begin{algorithmic}
\STATE {\bfseries Input:} $\tau^\prime$; {\bfseries Initialize:}  $Z \leftarrow {\bm 0}_{d\times d}$; $\xi \leftarrow {\bm 0}_{d\times 1}$;
$n \leftarrow 0$
\FOR{$t=1$ {\bfseries to} $\tau^\prime$}
\STATE Select $a_t$ uniformly at random in $\mathcal{A}_t$
\STATE Update $n \leftarrow n \,+$ the total number of non-missing entries in $x_{a}$ for all $a\in\mathcal{A}$
\STATE Update 
 $Z \leftarrow Z+ \sum_{a\in\mathcal{A}} x_{a,t}x_{a,t}^\top$, $\xi \leftarrow \xi + \sum_{a\in \mathcal{A}} x_{a,t}$
\ENDFOR
\FOR{$t =\tau^\prime+1$ {\bfseries to} $T$}
\STATE Update $n \leftarrow n \,+$ the total number of non-missing entries in $x_{a}$ for all $a\in\mathcal{A}$
\STATE Update 
 $Z \leftarrow Z+ \sum_{a\in\mathcal{A}} x_{a,t}x_{a,t}^\top$, $\xi \leftarrow \xi + \sum_{a\in \mathcal{A}} x_{a,t}$
\STATE Estimate parameters:
\STATE $\widehat{p} \leftarrow \frac{\max\{1, n \}}{t d K}$;  $\widehat{\nu}\leftarrow \frac{1}{t K \widehat{p}}\xi$; $\widehat{\Sigma}\leftarrow \frac{1}{t K} Z \circ \left(\frac{\widehat{p}-1}{\widehat{p}^2} I_{d\times d}+ \frac{1}{\widehat{p}^2} {\bm 1}_{d\times d} \right) - \widehat{\nu} \widehat{\nu}^\top$
 \STATE Estimate features $\widehat{z}_{a,t}\leftarrow[1;\overline{x}(\widehat{\nu},\widehat{\Sigma},x_{a,t})]$ for $a\in\mathcal{A}_t$.
 \STATE Select $a_t=\arg\max_{a\in \mathcal{A}_t}\max_{\theta\in\mathcal{C}_{t-1}}\langle \widehat{z}_{a,t},\theta\rangle$
\STATE Observe reward $y_t$.
\ENDFOR
\end{algorithmic}
\end{algorithm}

We first define some notations in Algorithm~\ref{alg:alg2}. Let $\lambda=Cd\log(KT)$,
$V_t= \lambda I_{d+1}+\sum_{s=\tau^\prime+1}^t \widehat{z}_{a_s,s}\widehat{z}_{a_s,s}^\top$, and 
$\widehat{\theta}_t=V_t^{-1} \sum_{s=\tau^\prime+1}^{t}\widehat{z}_{a_s,s}^\top y_s$  for some sufficiently large constant $C>0$. We note that $\widehat{z}_{a_s,s}$ only for $s>\tau^\prime$ are used in $V_t$ and $\widehat{\theta}_t$. We also define the confidence set for estimating a latent parameter $\overline{\theta}$ as
$$\mathcal{C}_t=\{\theta\in \mathbb{R}^{d+1}:\|\widehat{\theta}_t-\theta\|_{V_t}\le \sqrt{(d+1)\log((1+t)T)}+\lambda^{1/2} \}.$$ 

In Algorithm~\ref{alg:alg1}, $\widehat{z}_{a_s,s}$ in $V_t$ and $\widehat{\theta}_t$  is required to be updated after time step $s$ because estimated features in early stages may not be accurate. However, by introducing an explicit exploration phase for the first $\tau^\prime$ time steps, Algorithm~\ref{alg:alg2} does not use the early estimated features for $V_t$ and $\widehat{\theta}_t$. Therefore, in Algorithm~\ref{alg:alg2}, $\widehat{z}_{a_s,s}$ in $V_t$ and $\widehat{\theta}_t$ is computed at time $s$ and it will not be updated anymore in later time steps.

The algorithm introduces an \textit{explicit} exploration phase over $\tau^\prime$ time steps, which guarantees the well-estimated features after the phase. Therefore, it does not require updating estimated features for previously selected arms, which is the main bottleneck of the Algorithm~\ref{alg:alg1}.
 However, setting a proper $\tau^\prime$ for the exploration phase requires knowing an $\alpha$ that satisfies $\alpha\le p$, and the regret bound depends on $\alpha$. By setting $\tau^\prime=\lceil d(\log(T))^2/(K\alpha^4)\rceil$, we can show that the regret bound of the efficient algorithm is 
 $ R(T)=\tilde{O}\left(d\sqrt{T}+(d^2/p^{3/2})\sqrt{T/K}+d/(\alpha^4K)\right).$
 The last term in the regret bound is larger than that in the regret of Theorem~\ref{thm:regret} due to the additional exploration phase in the efficient algorithm.

Now we provide a proof for the regret bound.

\paragraph{Regret bound of Algorithm~\ref{alg:alg2}}
From Lemma~\ref{lem:max_r}, regret from $t=1$ to $\tau^\prime$ is bounded as $$\sum_{t=1}^{\tau^\prime}\mathbb{E} \left[ \overline{x}(\nu_f,\Sigma,x_{a^\star_t,t})^\top 
  \theta^\prime - \overline{x}(\nu_f,\Sigma,x_{a_t,t})^\top
  \theta^\prime\right]=O\left(\tau^\prime\sqrt{\log (K)} \right).$$
Then from Lemma~\ref{lem:sampletheta}, for any time $t> \tau^\prime$,  with probability at least $1-1/T$, we have $\|\Sigma-\widehat{\Sigma}_t\|_2=o(1)$, which incurs to have
\begin{gather*}
    \|\widehat{\Sigma}\|_2\le \|\widehat{\Sigma}-\Sigma\|_2+\|\Sigma\|_2=O\left(\|\Sigma\|_2\right) \quad\text{ and}
    \cr
    \|\widehat{\Sigma}^{-1}\|_2=\frac{1}{\sigma_d(\widehat{\Sigma})}\le\frac{1}{\sigma_d(\Sigma)-\|\widehat{\Sigma}-\Sigma\|_2}=O\left(\frac{1}{\sigma_d(\Sigma)}\right)=O\left(\|\Sigma^{-1}\|_2\right).
\end{gather*} 

Therefore, for all $t>\tau^\prime$, we have $$\|\widehat{z}_{a_t,t}\|_2=O(\sqrt{d\log(KT)}).$$
Since we can get the bound for the estimated features after $\tau^\prime$,
the algorithm utilizes the estimated features of chosen arms only after $\tau^\prime$ for getting $V_t$ and $\widehat{\theta}_t$. Then by following the proof steps in Theorem~\ref{thm:regret}, we can easily get
\begin{align*}
    R(T) &=\tilde{O}\left(d\sqrt{T}+\frac{d^2}{p^{3/2}}\sqrt{\frac{T}{K}}+\frac{d}{K\alpha^4}\right).
\end{align*}
\subsection{Real-world experiments }\label{app:real} 

\begin{figure}[ht]
    \centering
    {{\includegraphics[width=0.29\textwidth]{image/avazu.png} }}%
    \hspace{3mm}
    {{\includegraphics[width=0.29\textwidth]{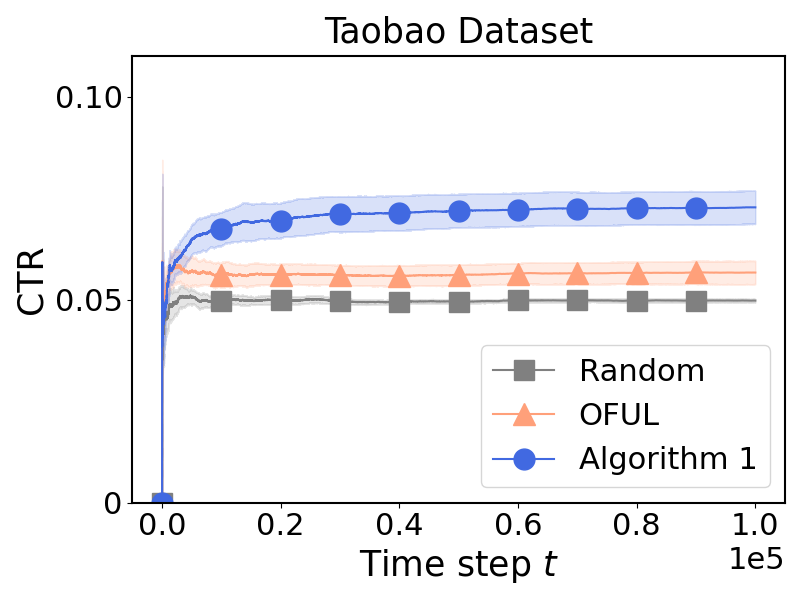}}}\hspace{3mm}
    {{\includegraphics[width=0.29\textwidth]{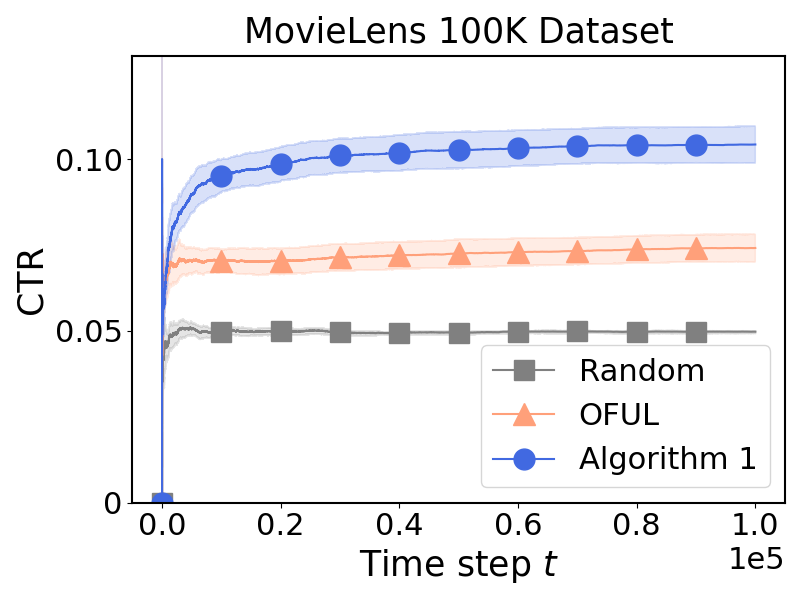}}}%
\caption{Click-through rate (CTR) for Algorithm~\ref{alg:alg1}, \texttt{OFUL}, and random policy for real-world datasets.} 
   \label{fig:real}%
\end{figure}

Here we present numerical results for real-world datasets. We compare our algorithm with \texttt{OFUL} and the random policy for click-through rate (CTR) at different time steps. 
For the comparison, we use Avazu CTR \citep{Avazu_CTR_dataset},  Taobao.com \citep{Taobao_dataset}, and MovieLens 100K \citep{harper2015movielens} 
datasets that contain CTR or rating information collected from advertising or movie recommendation systems. We use autoencoder models to preprocess feature information, constructing item feature vectors for each item. We set preprocessed feature dimension $d=32$, number of available items at each time, $K=20$, and time horizon $T=10^5$. We erase some entries in the preprocessed feature vectors using masking vectors with missing probability $1-p=0.1$. We examine average CTR for each algorithm over time steps $t\in[T]$. In Figure~\ref{fig:real}, we observe that our algorithm outperforms \texttt{OFUL} and the random policy for all the datasets
\textit{without generating both feature and noise vectors from Gaussian distributions externally}. In what follows, we describe the detailed experiment settings for real-world datasets. 

{\bfseries Avazu CTR.} This dataset contains mobile advertisement recommendation log data. Each data contains user, advertisement (ad), and click information. Each user-ad pair feature contains information about device type, site category, category of visited website, banner position, etc. For modeling reward payoffs, we use click information for each recommendation (user-ad); 0 for non-click and 1 for click. Using a pre-trained autoencoder model, we preprocess each user-ad feature vector to reduce dimension to $32$ (output of the encoder), which we refer to as an item feature. The autoencoder is trained using user-ad feature information without reward information.
Then, we erase each entry of the preprocessed item features with probability $0.1$ for modeling the missing data scenario. We then divide the items (user-ad pairs) into two sets according to whether the ads are clicked or not; randomly select 1000 items to construct each set. For each time, algorithms get an available item set, which is constructed by randomly selecting $K=20$ items from the two sets. In detail, the available item set consists of one selected from the clicked set and the others selected from the non-clicked set. Then, there must be one best item in the available item set. At each time, an algorithm selects an item from the available item set and get a reward according to the click information $0$ or $1$. Selecting an item can be represented as recommending an ad to a user.

{\bfseries Taobao.com.}
This dataset contains advertisement display/click log data on the website of Taobao.com. As in the Avazu dataset, each item (user-ad pair) contains user and ad information such as gender, age, consumption grade, brand, category, etc. Also there exist click data $1$ or $0$ for each recommendation. For each item feature, using a pre-trained autoencoder, we get a prepocessed feature vector with dimension $32$. The rest of the experiment setting is the same as the case for the Avazu CTR dataset. 

{\bfseries MovieLens 100K.} This dataset contains movie ratings from users collected through the MovieLens website. Each rating has an integer value from $1$ to $5$. For modeling binary reward payoffs, we treat rating $5$ as reward $1$ and otherwise $0$. Each user and movie data contain feature information such as age, gender, movie genre, etc. The rest of the experiment setting is the same as the experiment settings for the above datasets.

\end{document}